\documentclass[10pt,twocolumn,letterpaper]{article}

\usepackage[pagenumbers]{cvpr} 

\makeatletter
\@namedef{ver@everyshi.sty}{}
\makeatother

\usepackage{graphicx}
\usepackage{amsmath}
\usepackage{amssymb}
\usepackage{booktabs}
\usepackage{tabularx}
\usepackage{multirow}
\usepackage{units}
\usepackage{pifont}
\usepackage{tikz}
\usepackage{pgfplots}
\usepackage{relsize}
\usepackage[nolist]{acronym}

\graphicspath{{illustrations/}}

\newcommand{\cmark}{\ding{51}}%
\newcommand{\xmark}{\ding{55}}%

\DeclareMathOperator*{\trace}{trace}

\definecolor{dlrred}{RGB}{179,63,61}
\definecolor{dlrorange}{RGB}{234,184,24}
\definecolor{dlrblue}{RGB}{0,117,187}

\begin{acronym}
	\acro{ICP}[ICP]{Iterative Closest Point}
	\acro{PDF}[PDF]{probability density function}
	\acro{RMS}[RMS]{root mean square}
	\acro{6DoF}[6DoF]{six degrees of freedom}
	\acro{CNN}[CNN]{convolutional neural network}
\end{acronym}

\usepackage[pagebackref,breaklinks,colorlinks]{hyperref}

\usepackage[capitalize]{cleveref}
\crefname{section}{Sec.}{Secs.}
\Crefname{section}{Section}{Sections}
\Crefname{table}{Table}{Tables}
\crefname{table}{Tab.}{Tabs.}

\def\cvprPaperID{2568} 
\def\confName{CVPR}
\def\confYear{2022}

\begin{document}

\title{Iterative Corresponding Geometry: Fusing Region and Depth for\\ Highly Efficient 3D Tracking of Textureless Objects}

\author{Manuel Stoiber$^{1, 2}$ \qquad Martin Sundermeyer$^{1, 2}$ \qquad Rudolph Triebel$^{1, 2}$\\
$^{1}$	German Aerospace Center (DLR) \qquad $^{2}$ Technical University of Munich (TUM)\\
	{\tt\small\{firstname.lastname\}@dlr.de}
}
\maketitle


\begin{abstract}
Tracking objects in 3D space and predicting their \acs{6DoF} pose is an essential task in computer vision.
State-of-the-art approaches often rely on object texture to tackle this problem.
However, while they achieve impressive results, many objects do not contain sufficient texture, violating the main underlying assumption.
In the following, we thus propose \textit{ICG}, a novel probabilistic tracker that fuses region and depth information and only requires the object geometry.
Our method deploys correspondence lines and points to iteratively refine the pose.
We also implement robust occlusion handling to improve performance in real-world settings.
Experiments on the \textit{YCB-Video}, \textit{OPT}, and Choi datasets demonstrate that, even for textured objects, our approach outperforms the current state of the art with respect to accuracy and robustness.
At the same time, \textit{ICG} shows fast convergence and outstanding efficiency, requiring only $1.3\,\unit{ms}$ per frame on a single CPU core.
Finally, we analyze the influence of individual components and discuss our performance compared to deep learning-based methods.
The source code of our tracker is publicly available\footnote{\url{https://github.com/DLR-RM/3DObjectTracking}}.
\end{abstract}

\section{Introduction}\label{sec:in}
For many applications in robotic manipulation and augmented reality, it is essential to know the \ac{6DoF} pose of relevant objects.
To provide this information at high frequency, 3D object tracking is used.
The goal is to estimate an object's position and orientation from consecutive image frames given its 3D model.
In real-world applications, occlusions, motion blur, background clutter, textureless surfaces, object symmetries, and real-time requirements remain difficult problems.
Over the years many approaches have been developed \cite{Lepetit2005, Yilmaz2006}.
They can be differentiated by the use of keypoints, edges, direct optimization, deep learning, object regions, and depth images.

While methods based on keypoints \cite{Rublee2011, Ozuysal2006, Rosten2005, Skrypnyk2004, Vacchetti2004}, edges \cite{Seo2014, Comport2006, Drummond2002b, Harris1990}, and direct optimization \cite{Seo2016, Crivellaro2014, Benhimane2004, Lucas1981}
were very popular in the past, multiple drawbacks exist.
Both keypoints and direct optimization are not suitable for textureless objects.
Edge-based methods, on the other hand, typically struggle with background clutter and object texture.
Further problems emerge from reflections and motion blur, which change the appearance of both texture and edges.
To overcome those issues, data-driven techniques that use \acp{CNN} have been proposed \cite{Deng2021, Wen2020, Li2018, Wang2020}.
While most of those methods require significant computational resources and a detailed 3D model, they achieve promising results.
For the tracking of textureless objects in cluttered environments, region-based techniques have also become very popular \cite{Stoiber2021, Zhong2020, Tjaden2018, Prisacariu2012}.
Furthermore, the emergence of consumer depth sensors has enabled additional trackers that do not rely on texture \cite{Cifuentes2017, Issac2016, Schmidt2015, Wuethrich2013, Newcombe2011}.
Finally, while all those methods can be used independently, many approaches demonstrated the benefits of combining different techniques \cite{Ren2017, Kehl2017, Tan2017, Krainin2011}.%
\begin{figure}[t]
	\centering
	\includegraphics[width=\linewidth]{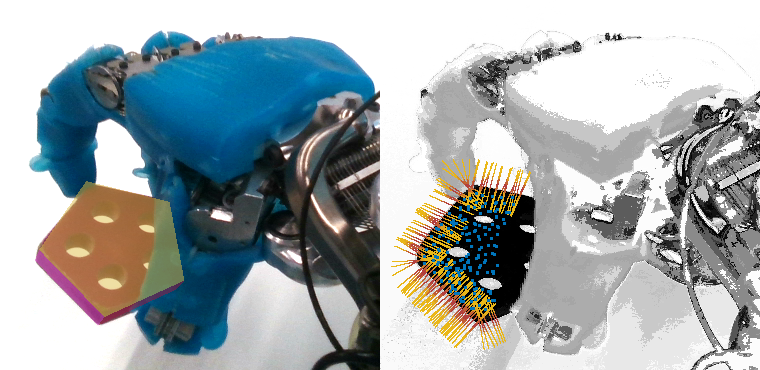}
	\caption{
		Tracking of a pentagon object for robotic manipulation.
		The image on the left shows an overlay of the object model for the predicted pose.
		On the right, probabilities that a pixel belongs to the background are encoded in a grayscale image.
		Correspondence lines are shown in yellow, with high probabilities indicated in red.
		Projected correspondence points are illustrated in blue.
	}\label{fig:in20}
\end{figure}%

In the past, it was shown that a combination of region and depth has great potential for the tracking of textureless objects \cite{Ren2017, Kehl2017}.
However, while region-based techniques improved greatly with respect to efficiency and quality \cite{Stoiber2021, Stoiber2020b}, no recent combined approach exists.
In the following, we thus build on current developments and propose \textit{ICG}, a highly efficient method that fuses geometric information from region-based correspondence lines and depth-based correspondence points.
An illustration of the used correspondences is shown in \cref{fig:in20}.
In a detailed evaluation on three different datasets, our method demonstrates state-of-the-art performance compared to both classical and deep learning-based techniques.
In addition, given that only few such comparisons were conducted in the past, we are able to gain new insights into the current state of deep learning-based object tracking and pose estimation.

\section{Related Work}\label{sec:r}
In the following, we provide an overview on region-, depth-, and deep learning-based techniques.
Region-based methods typically use color statistics to model the probability that a pixel belongs to the object or to the background.
The object pose is then optimized to best explain the segmentation of the image.
While early approaches treated segmentation and optimization separately \cite{Schmaltz2012, Brox2010, Rosenhahn2007}, subsequent work \cite{Dambreville2008} combined the two steps.
Later, the pixel-wise posterior membership of \cite{Bibby2008} was used to develop \textit{PWP3D} \cite{Prisacariu2012}.
Based on this method, multiple combined approaches that incorporate depth information \cite{Ren2017, Kehl2017}, edges \cite{Sun2021, Li2021}, inertial measurements \cite{Prisacariu2015}, or use direct optimization \cite{Liu2020, Zhong2019, Liu2021} were developed.
In addition, it was suggested to localize the probabilistic segmentation model \cite{Hexner2016, Tjaden2018, Zhong2020}.
Different optimization techniques such as particle filters \cite{Zhao2014}, Levenberg Marquardt \cite{Prisacariu2015}, Gauss Newton \cite{Tjaden2018}, or Newton with Tikhonov regularization \cite{Stoiber2020b, Stoiber2021} were also proposed.
Finally, starting from the ideas of \cite{Kehl2017}, the efficiency problem of region-based methods was addressed with the development of the sparse tracker \textit{SRT3D} \cite{Stoiber2020b, Stoiber2021}.

Depth-based methods try to minimize the distance between the surface of a 3D model and measurements from a depth camera.
Often, approaches based on the \textit{\ac{ICP}} framework \cite{Chen1992, Besl1992} are used.
While many variants exist \cite{Rusinkiewicz2001, Pomerleau2015}, all algorithms iteratively establish correspondences and minimize a respective error function.
For tracking, projective data association \cite{Blais1995} and the point-to-plane error metric \cite{Chen1992} are very common \cite{Newcombe2011, Pauwels2014, Tan2017, Kehl2017}.
Apart from correspondence points and \textit{\ac{ICP}}, methods that utilize signed distance functions are often used \cite{Fitzgibbon2003, Ren2012, Schmidt2015, Ren2017}.
In addition, approaches that employ particle filters \cite{Wuethrich2013, Choi2013, Krull2015, Cifuentes2017} or robust Gaussian filters \cite{Issac2016} instead of gradient-based optimization are also very popular.

While deep learning has proven highly successful for \ac{6DoF} pose estimation \cite{Labbe2020, He2021, Wang2019b, Xiang2018, Sundermeyer2018}, pure tracking methods were only recently proposed.
Many approaches are inspired by pose refinement and predict the relative pose between object renderings and subsequent images \cite{Li2018, Manhardt2018, Wen2020}. 
In addition, \textit{PoseRBPF} \cite{Deng2021} uses a Rao-Blackwellized particle filter on pose-representative latent codes \cite{Sundermeyer2018} while \textit{6-Pack} \cite{Wang2020} tracks anchor-based keypoints.


\section{Probabilistic Model}\label{sec:p}
In this section, mathematical concepts and the used notation are introduced.
This is followed by an explanation of the sparse viewpoint model.
Finally, the \acp{PDF} for region and depth are derived.

\subsection{Preliminaries}\label{ssec:p0}
In this work, we use $\pmb{X} = \begin{bmatrix} X& Y& Z\end{bmatrix}^\top\in \mathbb{R}^3$ and the homogeneous form $\pmb{\widetilde{X}} = \begin{bmatrix} X& Y& Z& 1\end{bmatrix}^\top$ to describe 3D model points.
Image coordinates $\pmb{x} = \begin{bmatrix} x& y\end{bmatrix}^\top \in \mathbb{R}^2$ are employed to access color values $\pmb{y} = \pmb{I}_\textrm{c}(\pmb{x})$ and depth values $d_Z = I_\textrm{d}(\pmb{x})$ from the respective color and depth images.
With the pinhole camera model, a 3D model point is projected into an undistorted image as follows
\begin{equation}\label{eq:p01}
	\renewcommand\arraystretch{1.2}
	\pmb{x} = \pmb{\pi}(\pmb{X}) = 
	\begin{bmatrix}
		\frac{X}{Z}f_x + p_x\\
		\frac{Y}{Z}f_y + p_y
	\end{bmatrix},
\end{equation}
where $f_x$ and $f_y$ are the focal lengths and $p_x$ and $p_y$ are the coordinates of the principal point.

To describe the relative pose between two reference frames $\textrm{A}$ and $\textrm{B}$, the homogeneous matrix ${}_A\pmb{T}_B{}\in \mathbb{SE}(3)$ is used.
It transforms 3D model points as follows
\begin{equation} \label{eq:p03}
	_\textrm{A}\pmb{\widetilde{X}} = {}_\textrm{A}\pmb{T}_\textrm{B}{}_\textrm{B}\pmb{\widetilde{X}} =
	\begin{bmatrix}
		_\textrm{A}\pmb{R}_\textrm{B} & _\textrm{A}\pmb{t}_\textrm{B} \\ \pmb{0} & 1
	\end{bmatrix}
	{}_\textrm{B}\pmb{\widetilde{X}},
\end{equation}
with $_\textrm{A}\pmb{\widetilde{X}}$ and $_\textrm{B}\pmb{\widetilde{X}}$ a point written in the coordinate frames $\textrm{A}$ and $\textrm{B}$.
The rotation matrix $_\textrm{A}\pmb{R}_\textrm{B} \in \mathbb{SO}(3)$ and the translation vector $_\textrm{A}\pmb{t}_\textrm{B} \in \mathbb{R}^3$ define the transformation from $\textrm{B}$ to $\textrm{A}$.
In this work, $\textrm{M}$, $\textrm{C}$, and $\textrm{D}$ will be used to denote the model, the color camera, and the depth camera frames, respectively.

For small variations of the pose in the model reference frame $\textrm{M}$, we use the following minimal representation
\begin{equation} \label{eq:p04}
	_\textrm{M}\pmb{\widetilde{X}}(\pmb{\theta}) =
	{}_\textrm{M}\pmb{T}(\pmb{\theta}){}_\textrm{M}\pmb{\widetilde{X}} =
	\begin{bmatrix}
		\pmb{I} + [\pmb{\theta}_\textrm{r}]_\times & \pmb{\theta}_\textrm{t} \\ \pmb{0} & 1
	\end{bmatrix}
	{}_\textrm{M}\pmb{\widetilde{X}},
\end{equation}
where $[\pmb{\theta}_\textrm{r}]_\times$ is the skew-symmetric matrix of $\pmb{\theta}_\textrm{r}$.
The vectors $\pmb{\theta}_\textrm{r}\in \mathbb{R}^3$ and $\pmb{\theta}_\textrm{t}\in \mathbb{R}^3$ are the rotational and translational components of the full variation vector $\pmb{\theta}^\top = \begin{bmatrix} \pmb{\theta}_\textrm{r}^\top & \pmb{\theta}_\textrm{t}^\top \end{bmatrix}$.

\subsection{Sparse Viewpoint Model}\label{ssec:p1}
To ensure efficiency and avoid the rendering of the 3D model during tracking, we represent the geometry using a sparse viewpoint model \cite{Stoiber2021}.
In the generation process, the object is rendered from a large number of virtual cameras that are placed on the vertices of a geodesic grid all around the object.
Similar to \cite{Tan2017}, image coordinates are then randomly sampled on the contour and surface of the rendered silhouette.
For each coordinate, both the 3D point ${}_\textrm{M}\pmb{X}$ and the 3D normal vector ${}_\textrm{M}\pmb{N}$ are reconstructed.
Together with the orientation that points from the camera to the model center, those vectors are stored for each viewpoint.
Given a pose estimate, obtaining the contour and surface representation reduces to a search for the closest orientation vector.

\subsection{Region Modality}\label{ssec:p2}
\begin{figure}[t]
	\centering
	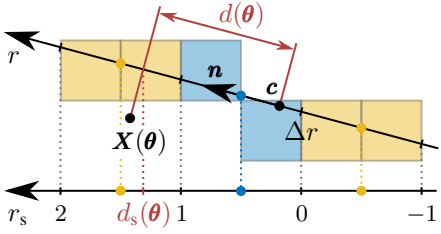
	\caption{
		Projection of a correspondence line from image space into scale space.
		The correspondence line is defined by a center $\pmb{c}$ and a normal vector $\pmb{n}$.
		Pixels along the correspondence line are combined into segments, which are illustrated in blue and yellow.
		The number of pixels per segment is specified by the scale $s = 2$.
		In the visualization, the pose-dependent 3D contour point $\pmb{X}(\pmb{\theta})$, which is associated with the correspondence line, is also shown.
		It is used to compute the distance $d(\pmb{\theta})$ from the estimated contour to the center $\pmb{c}$.
		The transformation from $r$ into the scale space $r_\textrm{s}$ is indicated by dotted vertical lines.
		Note that the space is shifted by $\Delta r$ and scaled according to $s$, as well as the line angle.
	}\label{fig:p20}
\end{figure}

In the following, we adopt the region-based approach of \textit{SRT3D} \cite{Stoiber2020b,Stoiber2021} and modify it to incorporate a user-defined uncertainty.
In general, \textit{SRT3D} considers region information sparsely along so-called correspondence lines $\pmb{l}$, which cross the estimated object contour.
Similar to the image function $\pmb{I}_\textrm{c}$, correspondence lines map coordinates $r \in \mathbb{R}$ to color values $\pmb{y} = \pmb{l}(r)$.
Each line is thereby defined by a center $\pmb{c} \in \mathbb{R}^2$ and a normal vector $\pmb{n} \in \mathbb{R}^2$ in image space.
Both vectors are calculated by projecting a 3D contour point $\pmb{X}$ and an associated 3D normal vector $\pmb{N}$ from the sparse viewpoint model into the image to establish a correspondence.
In order to make correspondence lines more efficient, \textit{SRT3D} introduces a scale-space formulation that combines multiple pixel values $\pmb{y}$ into segments $\pmb{s}$.
The number of pixels is thereby defined by the scale $s\in \mathbb{N}^+$.
In addition, line coordinates $r$ are scaled and shifted to make correspondence lines independent of their orientation and sub-pixel location.
An illustration is shown in \cref{fig:p20}.

Like in most region-based methods, color statistics are used to differentiate between foreground and background.
The probabilities $p(\pmb{y}\mid m_\textrm{f})$ and $p(\pmb{y}\mid m_\textrm{b})$ are approximated using normalized color histograms.
They describe the likelihood that pixel colors $\pmb{y}$ are part of the foreground or background model $m_\textrm{f}$ or $m_\textrm{b}$.
Based on those probabilities, segment-wise posteriors are calculated as
\begin{equation} \label{eq:p22}
	p_{\textrm{s}i}(r_\textrm{s}) = \frac{\prod\limits_{\pmb{y} \in \pmb{s}} p(\pmb{y}\mid m_i)} {\prod\limits_{\pmb{y} \in  \pmb{s}} p(\pmb{y} \mid  m_\textrm{f}) + \prod\limits_{\pmb{y} \in  \pmb{s}} p(\pmb{y} \mid  m_\textrm{b})} , \quad i\in\{\textrm{f}, \textrm{b}\},
\end{equation}
where the segment $\pmb{s}$ is defined by the coordinate $r_\textrm{s}$.
The value describes the probability that a specific segment belongs to the foreground or background.

In addition to those measurements, theoretical probabilities that depend on the location of the object contour are developed.
They are modeled by smoothed step functions 
\begin{align}\label{eq:p23}
	h_\textrm{f}(x) &= \frac{1}{2} - \alpha_\textrm{h}\tanh\bigg(\frac{x}{2s_\textrm{h}}\bigg),\\\label{eq:p24}
	h_\textrm{b}(x) &= \frac{1}{2} + \alpha_\textrm{h}\tanh\bigg(\frac{x}{2s_\textrm{h}}\bigg),
\end{align}
with the amplitude parameter $\alpha_\textrm{h} \in [0,0.5]$ and the slope parameter $s_\textrm{h} \in \mathbb{R}^+$.
Using the variated 3D model point
\begin{equation} \label{eq:p20}
	{}_\textrm{C}\pmb{\widetilde{X}}(\pmb{\theta}) = {}_\textrm{C}\pmb{T}_\textrm{M}{}_\textrm{M}\pmb{T}(\pmb{\theta}){}_\textrm{M}\pmb{\widetilde{X}},
\end{equation}
the distance from the estimated contour to the correspondence line center $\pmb{c}$ is approximated in scale space as follows
\begin{equation} \label{eq:p21}
	d_\textrm{s}(\pmb{\theta}) = \bigg(\pmb{n}^\top\Big(\pmb{\pi}\big({}_\textrm{C}\pmb{X}(\pmb{\theta})\big) - \pmb{c}\Big)-\Delta r\bigg) \frac{\bar{n}}{s},
\end{equation}
where $\bar{n} = \lVert\pmb{n}\rVert_\mathrm{max}$ projects to the closest horizontal or vertical image coordinate, and $\Delta r \in \mathbb{R}$ is an offset to a defined pixel location.
An illustration of the transformation is shown in \cref{fig:p20}.
Finally, based on those functions, \textit{SRT3D} estimates the \ac{PDF} for the scaled contour distance as
\begin{equation}\label{eq:p25}
	p(d_\textrm{s}(\pmb{\theta})\mid \omega_\textrm{s},\pmb{l}) \propto \prod_{r_\textrm{s}\in\omega_\textrm{s}}
	\sum_{i \in \{\textrm{f}, \textrm{b}\}}
	h_i\big(r_\textrm{s}-d_\textrm{s}(\pmb{\theta})\big)p_{\textrm{s}i}(r_\textrm{s}),
\end{equation}
with $\omega_\textrm{s}$ the considered correspondence line domain.

Note that this \ac{PDF} is defined in scale space.
Thanks to the proof in \cite{Stoiber2020b}, we know that, under certain conditions, the variance of the \ac{PDF} is equal to the slope parameter $s_\textrm{h}$ defined for the smoothed step functions $h_\textrm{f}$ and $h_\textrm{b}$.
Given this variance, the expected unscaled variance in units of pixels is $\sigma^2 = \nicefrac{s_\textrm{h} s^2}{\bar{n}^2}$.
In contrast to previous work, we want correspondence lines to be independent of scale and slope parameters.
This has the advantage that for all correspondence lines, the variance is defined in the same unit of pixels.
In addition, we want to define our confidence in the region modality.
Introducing the user-defined standard deviation $\sigma_\textrm{r}$, the \ac{PDF} in \cref{eq:p25} is thus scaled as follows
\begin{equation}\label{eq:t22}
p(\pmb{\theta}\mid\omega_{\textrm{s}},\pmb{l}) \propto p(d_{\textrm{s}}(\pmb{\theta})\mid\omega_{\textrm{s}},\pmb{l})^\mathlarger{\frac{s_\textrm{h}s^2}{\sigma_\textrm{r}^2\bar{n}^2}}.
\end{equation}
The formulation helps to fuse the region modality with other information, given a defined uncertainty.
Also, as shown in \cref{sec:a2}, it improves results compared to \textit{SRT3D}.

\subsection{Depth Modality}\label{ssec:p3}
Based on \textit{\ac{ICP}} \cite{Chen1992, Besl1992}, the depth modality starts with a search for correspondence points.
Similar to projective data association \cite{Blais1995}, a 3D surface point $\pmb{X}$ from the sparse viewpoint model is first projected into the depth image.
Given a user-defined radius and stride, multiple 3D points within a quadratic area are then reconstructed.
Finally, a correspondence point $\pmb{P}\in \mathbb{R}^3$ is selected as the point closest to the point $\pmb{X}$.
Correspondences with a distance bigger than a threshold are rejected.
Note that techniques such as normal shooting \cite{Chen1992, Gagnon1994}, and rejection strategies based on the median distance \cite{Diebel2004}, the best percentage \cite{Pulli1999}, and the compatibility of normal vectors \cite{Zhang1994} were also tested.
However, in the end, this simple procedure worked best.

For the probabilistic model, we formulate a normal distribution that uses the point-to-plane error metric \cite{Chen1992}.
The distance between the 3D surface point $\pmb{X}$ and correspondence point $\pmb{P}$ is thereby calculated along the associated normal vector $\pmb{N}$.
Given the correspondence point $\pmb{P}$, the probability for the pose variation vector $\pmb{\theta}$ is written as
\begin{equation}\label{eq:p30}
	p(\pmb{\theta}\mid \pmb{P}) \propto \exp\bigg(\hspace{-0.1cm}-\frac{1}{2 d_\textrm{Z}^2 \sigma_\textrm{d}^2 }\Big({}_\textrm{M}\pmb{N}^\top\big({}_\textrm{M}\pmb{X} - {}_\textrm{M}\pmb{P}(\pmb{\theta})\big)\Big)^{\hspace{-0.05cm}2}\bigg)
\end{equation}
with
\begin{equation}\label{eq:p31}
	{}_\textrm{M}\pmb{\widetilde{P}}(\pmb{\theta}) = {}_\textrm{M}\pmb{T}(-\pmb{\theta}) {}_\textrm{M}\pmb{T}_\textrm{D} {}_\textrm{D}\pmb{\widetilde{P}}.
\end{equation}
Note that the user-defined standard deviation $\sigma_\textrm{d}$ is scaled by the depth value $d_\textrm{Z}$ of the correspondence point $\pmb{P}$.
The scaling takes into account that the number and quality of depth measurements decreases with the distance to the camera.
In addition, it also ensures compatibility with the uncertainty of the region modality, which increases with the camera distance.
In \cref{eq:p30}, we variate the correspondence point $\pmb{P}$ instead of the model.
This has the advantage that the normal vector remains fixed, and only one vector has to be variated.
Based on the derived \acp{PDF} for region and depth, we can now optimize for the pose that best explains the data.

\section{Optimization}\label{sec:o}
In the following, we first introduce the Newton method with Tikhonov regularization that is used to maximize the probability.
Subsequently, we define the gradient vector and the Hessian matrix that are required in this optimization.

\subsection{Regularized Newton Method}\label{ssec:o0}
\begin{figure}[t]
	\centering
\begingroup%
  \makeatletter%
  \providecommand\color[2][]{%
    \errmessage{(Inkscape) Color is used for the text in Inkscape, but the package 'color.sty' is not loaded}%
    \renewcommand\color[2][]{}%
  }%
  \providecommand\transparent[1]{%
    \errmessage{(Inkscape) Transparency is used (non-zero) for the text in Inkscape, but the package 'transparent.sty' is not loaded}%
    \renewcommand\transparent[1]{}%
  }%
  \providecommand\rotatebox[2]{#2}%
  \newcommand*\fsize{\dimexpr\f@size pt\relax}%
  \newcommand*\lineheight[1]{\fontsize{\fsize}{#1\fsize}\selectfont}%
  \ifx\svgwidth\undefined%
    \setlength{\unitlength}{184.2519685bp}%
    \ifx\svgscale\undefined%
      \relax%
    \else%
      \setlength{\unitlength}{\unitlength * \real{\svgscale}}%
    \fi%
  \else%
    \setlength{\unitlength}{\svgwidth}%
  \fi%
  \global\let\svgwidth\undefined%
  \global\let\svgscale\undefined%
  \makeatother%
  \begin{picture}(1,0.50769231)%
    \lineheight{1}%
    \setlength\tabcolsep{0pt}%
    \put(0,0){\includegraphics[width=\unitlength,page=1]{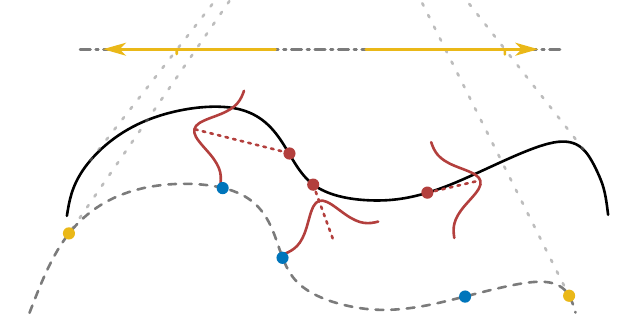}}%
    \put(0.19942219,0.10494573){\color[rgb]{0.48235294,0.48235294,0.48235294}\makebox(0,0)[lt]{\lineheight{0}\smash{\begin{tabular}[t]{l}\small Model\end{tabular}}}}%
    \put(0,0){\includegraphics[width=\unitlength,page=2]{corresponding_geometry.pdf}}%
    \put(0.49101031,0.26937819){\color[rgb]{0,0,0}\makebox(0,0)[lt]{\lineheight{0}\smash{\begin{tabular}[t]{l}\small Object\end{tabular}}}}%
    \put(0.38125097,0.46994488){\color[rgb]{0.48235294,0.48235294,0.48235294}\makebox(0,0)[lt]{\lineheight{0}\smash{\begin{tabular}[t]{l}\small Image Plane\end{tabular}}}}%
  \end{picture}%
\endgroup%

	\caption{
		Optimization of the corresponding geometry.
		For blue surface points and normal vectors, the probability given a red correspondence point is illustrated by a normal distribution.
		For yellow correspondence lines, the discrete distribution is illustrated.
		The location of 3D contour points is thereby projected to the image plane.
		During optimization, the joint probability is maximized.
	}\label{fig:o00}
\end{figure}

Assuming that measurements from both modalities are independent, the joint probability function is written as
\begin{equation} \label{eq:o01}
	p(\pmb{\theta}\mid \pmb{\mathcal{D}}) = \prod_{i = 0}^{n_\textrm{r}}
	p(\pmb{\theta}\mid\omega_{\textrm{s}i},\pmb{l}_i)
	\prod_{i = 0}^{n_\textrm{d}} p(\pmb{\theta}\mid \pmb{P}_i),
\end{equation}
with $\pmb{\mathcal{D}}$ the considered data and $n_\textrm{r}$ and $n_\textrm{d}$ the number of used correspondence lines and correspondence points, respectively.
The joint optimization of correspondence lines and correspondence points is visualized in \cref{fig:o00}.

To maximize the probability, multiple iterations, where we calculate the variation vector $\pmb{\hat{\theta}}$ and update the object pose, are performed.
In each iteration, the Newton method with Tikhonov regularization is used as follows
\begin{equation} \label{eq:o00}
	\pmb{\hat{\theta}} = \bigg(-\pmb{H} + 
	\begin{bmatrix}
		\lambda_\textrm{r} \pmb{I}_3 & \pmb{0}\\
		\pmb{0} & \lambda_\textrm{t} \pmb{I}_3
	\end{bmatrix}
	\bigg)^{-1}\pmb{g},
\end{equation}
where $\pmb{g}$ is the gradient vector, $\pmb{H}$ the Hessian matrix, and $\lambda_\textrm{r}$ and $\lambda_\textrm{t}$ are the rotational and translational regularization parameters.
The gradient vector and the Hessian matrix are thereby defined as the first- and second-order derivatives of the logarithm of the joint probability function in \cref{eq:o01}.

The big advantage of the Newton formulation is that, with the Hessian matrix, uncertainty is considered in all dimensions.
This means that, in addition to weighting the two modalities with $\sigma_\textrm{r}$ and $\sigma_\textrm{d}$, it also takes into account how well each correspondence constrains the different directions.
In addition, Tikhonov regularization acts as a prior probability that controls how much we believe in the previous pose.
For directions with little information, this regularization helps to keep the optimization stable.

Finally, given the knowledge that $\pmb{\hat{\theta}}_\textrm{r}$ corresponds to the rotation vector of the axis-angle representation, we are able to update the pose using the exponential map as follows 
\begin{equation} \label{eq:o02}
	_\textrm{A}\pmb{T}_\textrm{M}^+ =
	{}_\textrm{A}\pmb{T}_\textrm{M}
	\begin{bmatrix}
		\exp([\pmb{\hat{\theta}}_\textrm{r}]_\times) & \pmb{\hat{\theta}}_\textrm{t} \\ \pmb{0} & 1
	\end{bmatrix}
 	, \quad \textrm{A}\in\{\textrm{C}, \textrm{D}\}.
\end{equation}
Note that, typically, either the pose with respect to the color or the depth camera is calculated using \cref{eq:o02}.
The other is then updated using the known relative transformation.

\subsection{Gradient and Hessian}\label{ssec:o1}
Because the logarithm is applied in the calculation of the gradient vector and the Hessian matrix, products turn into summations and, based on \cref{eq:o01}, we can write
\begin{align}\label{eq:o10}
	\pmb{g} &= \sum_{i=1}^{n_\textrm{r}}\pmb{g}_{\textrm{r}i}
	+ \sum_{i=1}^{n_\textrm{d}}\pmb{g}_{\textrm{d}i},\\\label{eq:o11}
	\pmb{H} &= \sum_{i=1}^{n_\textrm{r}}\pmb{H}_{\textrm{r}i}
	+ \sum_{i=1}^{n_\textrm{d}}\pmb{H}_{\textrm{d}i},
\end{align}
where $\pmb{g}_{\textrm{r}i}$ and $\pmb{H}_{\textrm{r}i}$ are calculated from individual correspondence lines of the region modality, and $\pmb{g}_{\textrm{d}i}$ and $\pmb{H}_{\textrm{d}i}$ are based on correspondence points from the depth modality.

For the region modality, we apply the chain rule to calculate gradient vectors and Hessian matrices as follows
\begin{align}\label{eq:o12}
	\pmb{g}_{\textrm{r}i} &= \frac{s_\textrm{h}s^2}{\sigma_\textrm{r}^2\bar{n}_i^2}
	\frac{\partial\ln\big(p(d_{\textrm{s}i}\mid\omega_{\textrm{s}i},\pmb{l}_i)\big)}{\partial d_{\textrm{s}i}}
	\frac{\partial d_{\textrm{s}i}}{\partial {}_\textrm{C}\pmb{X}_{i}}
	\frac{\partial {}_\textrm{C}\pmb{X}_{i}}{\partial \pmb{\theta}}
	\bigg\vert_{\pmb{\theta}=\pmb{0}},\\\label{eq:o13}
	\begin{split}
		\pmb{H}_{\textrm{r}i} &\approx \frac{s_\textrm{h}s^2}{\sigma_\textrm{r}^2\bar{n}_i^2}
		\frac{\partial^2\ln\big(p(d_{\textrm{s}i}\mid\omega_{\textrm{s}i},\pmb{l}_i)\big)}{\partial {d_{\textrm{s}i}}^2}
		\left(
		\frac{\partial d_{\textrm{s}i}}{\partial{}_\textrm{C}\pmb{X}_{i}}\frac{\partial {}_\textrm{C}\pmb{X}_{i}}{\partial \pmb{\theta}}
		\right)^\top\\
		&\hspace{1.25cm} \left(\frac{\partial d_{\textrm{s}i}}{\partial _\textrm{C}\pmb{X}_{i}}
		\frac{\partial _\textrm{C}\pmb{X}_{i}}{\partial \pmb{\theta}}\right)
		\bigg\vert_{\pmb{\theta}=\pmb{0}}.
	\end{split}
\hspace{-0.2cm}
\end{align}
Note that, similar to \cite{Stoiber2020b}, second-order partial derivatives for $d_{\textrm{s}i}$ and $_\textrm{C}\pmb{X}_{i}$ are neglected.
Using \cref{eq:p20,eq:p21}, the following first-order partial derivatives can be calculated
\begin{align}\label{eq:o14}
	\begin{split}
		\frac{\partial d_{\textrm{s}i}}{\partial _\textrm{C}\pmb{X}_{i}} &=
		\frac{\bar{n}_i}{s}
		\frac{1}{{} _\textrm{C}Z_{i}^2}
		\big[
		\begin{matrix}
			n_{xi} f_x {}_\textrm{C}Z_{i} & n_{yi} f_y {}_\textrm{C}Z_{i}
		\end{matrix}\\
		&\hspace{1.75cm}\begin{matrix}
			-n_{xi} f_x {}_\textrm{C}X_{i} - n_{yi} f_y {}_\textrm{C}Y_{i}
		\end{matrix}
		\big],
	\end{split}\\\label{eq:o15}
	\frac{\partial _\textrm{C}\pmb{X}_{i}}{\partial\pmb{\theta}} &=
	{}_\textrm{C}\pmb{R}_\textrm{M} 
	\begin{bmatrix}
		-[_\textrm{M}\pmb{X}_i]_\times & \pmb{I}_3
	\end{bmatrix}.
\end{align}

To estimate the first- and second-order partial derivatives for the posterior probability distribution $p(d_{\textrm{s}i}\mid\omega_{\textrm{s}i},\pmb{l}_i)$, we use the same techniques as in \cite{Stoiber2021} and differentiate between global and local optimization.
For global optimization, the \ac{PDF} is sampled over $d_{\textrm{s}i}$, and the mean $\mu_i$ and variance $\sigma_i^2$ are calculated to approximate a normal distribution.
Based on this normal distribution, derivatives are calculated as
\begin{align}\label{eq:o16}
	\frac{\partial\ln\big(p(d_{\textrm{s}i}\mid\omega_{\textrm{s}i},\pmb{l}_i)\big)}{\partial {d_{\textrm{s}i}}} &\approx -\frac{1}{\sigma_i^2}(d_{\textrm{s}i} - \mu_i),\\ \label{eq:o17}
	\frac{\partial^2\ln\big(p(d_{\textrm{s}i}\mid\omega_{\textrm{s}i},\pmb{l}_i)\big)}{\partial {d_{\textrm{s}i}}^2} &\approx -\frac{1}{\sigma_i^2}.
\end{align}
For local optimization, the two probability values corresponding to the discrete distances $d_{\textrm{s}i}^-$ and $d_{\textrm{s}i}^+$ closest to $d_{\textrm{s}i}$ are used to approximate first-order partial derivatives as
\begin{equation}\label{eq:o18}
	\frac{\partial\ln\big(p(d_{\textrm{s}i}\mid\omega_{\textrm{s}i},\pmb{l}_i)\big)}{\partial {d_{\textrm{s}i}}} \approx \frac{\alpha_\textrm{s}}{\sigma_i^2}\ln\bigg(\frac{p(d_{\textrm{s}i}^+\mid\omega_{\textrm{s}i},\pmb{l}_i}{p(d_{\textrm{s}i}^-\mid\omega_{\textrm{s}i},\pmb{l}_i)}\bigg),
\end{equation}
where $\alpha_\textrm{s}$ is a user-defined learning rate.
Second-order partial derivatives are again calculated according to \cref{eq:o17}.

Finally, for the depth modality, gradient vectors and Hessian matrices can be calculated using \cref{eq:p30,eq:p31}
\begin{align}\label{eq:o19}
	\pmb{g}_{\textrm{d}i} &= -\frac{1}{d_Z^2 \sigma_d^2}  {}_\textrm{M}\pmb{N}_i^\top
	\big({}_\textrm{M}\pmb{X}_i - {}_\textrm{M}\pmb{P}_i \big)
	\begin{bmatrix}
		{}_\textrm{M}\pmb{P}_i {\times}{}_\textrm{M}\pmb{N}_i \\
		{}_\textrm{M}\pmb{N}_i
	\end{bmatrix},\\\label{eq:o20}
	\pmb{H}_{\textrm{d}i} &= -\frac{1}{d_Z^2 \sigma_d^2}  
	\begin{bmatrix}
		{}_\textrm{M}\pmb{P}_i \times {}_\textrm{M}\pmb{N}_i \\
		{}_\textrm{M}\pmb{N}_i
	\end{bmatrix}
	\begin{bmatrix}
		{}_\textrm{M}\pmb{P}_i \times {}_\textrm{M}\pmb{N}_i \\
		{}_\textrm{M}\pmb{N}_i
	\end{bmatrix}^\top.
\end{align}

\section{Implementation}\label{sec:i}
The following section provides implementation details, discusses how \textit{ICG} can be used for pose refinement, and explains how occlusions are handled.
For our implementation, we built on the code of \textit{SRT3D} \cite{Stoiber2021}.
To generate the sparse viewpoint model, the object is rendered from $2562$ virtual cameras that are placed on a geodesic grid with a distance of $0.8\,\unit{m}$ to the object center.
For each view, contour and surface points are sampled and normal vectors are approximated.
For contour points, distances along the normal vector for which the foreground and background are not interrupted are also computed.
To ensure only one transition exists on a correspondence line, lines with at least one of the two distances below $3$ segments are rejected.

The two color histograms for foreground and background are discretized by 4096 equidistant bins.
In their calculation, the first 20 pixels from the line center are used.
During tracking, we update histograms once the final pose was computed using the online adaptation of \cite{Bibby2008} with a learning rate of $\alpha = 0.2$.
In addition to tracking, our algorithm can also be used for pose refinement.
In such cases, we initialize histograms at each iteration before correspondences are established.
Since we do not perform a continuous update, histograms do not show the same quality as for tracking.
Nevertheless, they still include useful information that helps the algorithm to converge.
Experiments that demonstrate the performance for pose refinement are provided in \cref{sec:a4}.

For the region modality, the probability distributions $p(d_{\textrm{s}i}\mid\omega_{\textrm{s}i},\pmb{l}_i)$ are evaluated at $12$ discrete distance values $d_{\textrm{s}i}\in\{-5.5,-4.5,\dots,5.5\}$.
In the calculation of each probability value, we use $8$ precomputed values for the smoothed step functions $h_\textrm{f}$ and $h_\textrm{b}$ that correspond to $x\in\{-3.5, -2.5,\dots,3.5\}$.
Also, we define the slope parameter $s_\textrm{h} = 0.5$, the amplitude parameter $\alpha_\textrm{h} = 0.43$, and the learning rate $\alpha_\textrm{s} = 1.3$.
To find correspondence points for the depth modality, image values on a quadratic grid with a stride of $5\,\unit{mm}$ and a radius equal to the correspondence threshold $r_\textrm{t}$ are considered.
Both parameter values are projected from meters into pixels based on the depth of the 3D surface point.
Correspondence points with a distance that is bigger than the threshold $r_\textrm{t}$ are rejected.
Valid correspondences are then used in the optimization with the regularization parameters $\lambda_\textrm{r} = 1000$ and $\lambda_\textrm{t} = 30000$.

To find the final pose, we conduct 4 iterations in which correspondences are established.
For each iteration, the standard deviations $\sigma_\textrm{r}$ and $\sigma_\textrm{d}$, the scale $s$, and the threshold $r_\textrm{t}$ can be adjusted.
This allows to define our confidence in the data and the range in which region and depth information is considered.
Many characteristics such as the resolution, depth image quality, or frame-to-frame pose difference depend on the sequence.
We thus adjust the parameters for every dataset and provide them in the evaluation section.
Finally, in each iteration, two optimization steps are conducted.
For the region modality, global optimization is used in the first and local optimization in the second.

\begin{figure}[t]
	\centering
\begingroup%
  \makeatletter%
  \providecommand\color[2][]{%
    \errmessage{(Inkscape) Color is used for the text in Inkscape, but the package 'color.sty' is not loaded}%
    \renewcommand\color[2][]{}%
  }%
  \providecommand\transparent[1]{%
    \errmessage{(Inkscape) Transparency is used (non-zero) for the text in Inkscape, but the package 'transparent.sty' is not loaded}%
    \renewcommand\transparent[1]{}%
  }%
  \providecommand\rotatebox[2]{#2}%
  \newcommand*\fsize{\dimexpr\f@size pt\relax}%
  \newcommand*\lineheight[1]{\fontsize{\fsize}{#1\fsize}\selectfont}%
  \ifx\svgwidth\undefined%
    \setlength{\unitlength}{184.2519685bp}%
    \ifx\svgscale\undefined%
      \relax%
    \else%
      \setlength{\unitlength}{\unitlength * \real{\svgscale}}%
    \fi%
  \else%
    \setlength{\unitlength}{\svgwidth}%
  \fi%
  \global\let\svgwidth\undefined%
  \global\let\svgscale\undefined%
  \makeatother%
  \begin{picture}(1,0.38461538)%
    \lineheight{1}%
    \setlength\tabcolsep{0pt}%
    \put(0,0){\includegraphics[width=\unitlength,page=1]{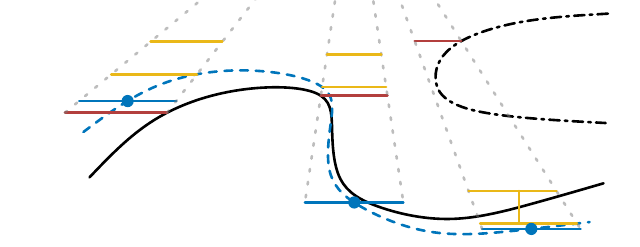}}%
    \put(0.772077,0.26008958){\makebox(0,0)[lt]{\lineheight{1.25}\smash{\begin{tabular}[t]{l}\small Occlusion\end{tabular}}}}%
    \put(0.27773118,0.13848314){\makebox(0,0)[lt]{\lineheight{1.25}\smash{\begin{tabular}[t]{l}\small Tracked\\\end{tabular}}}}%
    \put(0.2964636,0.08671312){\makebox(0,0)[lt]{\lineheight{1.25}\smash{\begin{tabular}[t]{l}\small Object\end{tabular}}}}%
    \put(0.36264064,0.28887282){\color[rgb]{0,0.45882353,0.73333333}\makebox(0,0)[lt]{\lineheight{1.25}\smash{\begin{tabular}[t]{l}\small Model\end{tabular}}}}%
    \put(0,0){\includegraphics[width=\unitlength,page=2]{occlusion_handling.pdf}}%
  \end{picture}%
\endgroup%

	\caption{
		Visualization of the occlusion handling strategy.
		For each blue model point, the considered region is defined by a blue line and dotted gray cone.
		The lower yellow line in each cone visualizes the depth value that is calculated from the model offset, while the upper yellow line adds the user-defined threshold.
		Red lines indicate the minimum depth measurements from the camera.
		For the right point, an occlusion is detected since the red depth measurement is smaller than the expected value in yellow.
	}\label{fig:i00}
\end{figure}

In many real-world cases, correspondence lines and correspondence points are subject to occlusion.
Based on measurements from the depth camera and estimated positions of 3D model points, occlusions can be detected.
First, the minimum depth based on $25$ depth image values within a quadratic region of $20\times20\,\unit{mm}$ is computed.
Similarly, during model generation, an offset between the depth of the sampled model point and the minimum depth within a quadratic region of $20\times20\,\unit{mm}$ is calculated.
Finally, we are able to reject correspondences for which the depth of the model point minus the precomputed offset and a user-defined threshold of $30\,\unit{mm}$ is smaller than the measured minimum depth.
Considering a region of values makes the technique robust to missing depth measurements and large local depth differences in the object surface.
In cases where depth images are not available, depth renderings can be used.
An illustration of the strategy is shown in \cref{fig:i00}.

\section{Evaluation}\label{sec:e}
In this section, we present an extensive evaluation of our approach on the \textit{YCB-Video} dataset \cite{Xiang2018}, the \textit{OPT} dataset \cite{Wu2017}, and the dataset of Choi \cite{Choi2013}.
We thereby evaluate the robustness, accuracy, and efficiency of \textit{ICG} in comparison to the state of the art.
In addition, we conduct an ablation study that demonstrates the importance of individual components.
Finally, we explain limitations of our approach.
Note that in \cref{sec:a1,sec:a2} we present further results for the region modality and the Choi dataset.
Also, in \cref{sec:a3,sec:a4}, we discuss how \textit{ICG} performs for pose refinement and how it compares to modern \ac{6DoF} pose estimation.
Qualitative results on the \textit{YCB-Video} dataset and in real-world applications can be found in the provided videos.

\subsection{YCB-Video Dataset}\label{ssec:e0}
The \textit{YCB-Video} dataset \cite{Xiang2018} contains $21$ \textit{YCB} objects \cite{Calli2015} and evaluates on $12$ sequences with a total of $2949$ key frames.
Because it includes additional training sequences, it is very popular with deep learning-based methods.
In the evaluation, the conventional and symmetric average distance errors $e_\textrm{ADD}$ and $e_\textrm{ADD\textnormal{-}S}$ \cite{Hinterstoisser2013} are calculated as
\begin{align}\label{eq:e00}
	e_\textrm{ADD} &= \frac{1}{n}\sum_{i=1}^n\big\lVert \big({}_\textrm{M}\pmb{\widetilde{X}}_i -  {}_\textrm{M}\pmb{T}_{\textrm{M}_\textrm{gt}}{}_\textrm{M}\pmb{\widetilde{X}}_i\big)_{3\times 1}\big\rVert_2,\\\label{eq:e01}
	e_\textrm{ADD\textnormal{-}S} &= \frac{1}{n}\sum_{i=1}^n \min_{j\in[n]} \big\lVert \big({}_\textrm{M}\pmb{\widetilde{X}}_i -  {}_\textrm{M}\pmb{T}_{\textrm{M}_\textrm{gt}}{}_\textrm{M}\pmb{\widetilde{X}}_j\big)_{3\times 1}\big\rVert_2,
\end{align}
where ${}_\textrm{M}\pmb{T}_{\textrm{M}_\textrm{gt}}$ is the difference between the ground-truth and the estimated model pose, $\pmb{X}_i$ is a vertex from the object mesh, and $n$ is the number of vertices.
Based on those error metrics for a single frame, \cite{Xiang2018} reports \textit{ADD} and \textit{ADD-S} area under curve scores.
They can be calculated as
\begin{equation}
	s_i = \frac{1}{m}\sum_{j=1}^m \max\Big(1 - \frac{e_{ij}}{e_\textrm{t}}, 0\Big),
\end{equation}
with $i \in \{\textrm{ADD}, \textrm{ADD\textnormal{-}S}\}$, the respective frame error $e_{ij}$, the number of frames $m$, and the threshold $e_\textrm{t} = 0.1\,\unit{m}$.

Results of the evaluation on the \textit{YCB-Video} dataset are shown in \cref{tab:e00}.
\begin{table*}
	\caption{
		Results on the \textit{YCB-Video} dataset \cite{Xiang2018} with \textit{ADD} and \textit{ADD-S} area under curve scores in percent.
		Except for \textit{PoseRBPF}\cite{Deng2021}, results are taken from \cite{Wen2020}.
		For \textit{DeepIM} \cite{Li2018}, the score over all frames was adjusted to be consistent with the evaluation of other methods.
		Objects with no conclusive geometry are indicated by a $^\star$ while objects with no or very little texture are marked by a $^\diamond$.
	}\label{tab:e00}

\centering
\scriptsize
\begin{tabularx}{\textwidth}{X *{7}{| >{\centering\arraybackslash}p{0.84cm}@{\hspace{0.0cm}} >{\centering\arraybackslash}p{0.84cm}}}
\hline
\noalign{\smallskip}
\multirow{2}{1cm}{\textbf{Approach}} & 
\multicolumn{2}{c|}{\multirow{2}{1.64cm}{\centering PoseCNN + ICP + DeepIM \cite{Xiang2018}}}& 
\multicolumn{2}{c|}{\multirow{2}{1.64cm}{\centering Wüthrich \cite{Wuethrich2013}}}&
\multicolumn{2}{c|}{\multirow{2}{1.64cm}{\centering RGF \cite{Issac2016}}}&
\multicolumn{2}{c|}{\multirow{2}{1.64cm}{\centering DeepIM \cite{Li2018}}}&
\multicolumn{2}{c|}{\multirow{2}{1.64cm}{\centering se(3)-TrackNet \cite{Wen2020}}}&
\multicolumn{2}{c|}{\multirow{2}{1.64cm}{\centering PoseRBPF + SDF \cite{Deng2021}}}&
\multicolumn{2}{c}{\multirow{2}{1.64cm}{\centering ICG (Ours)}}\\
&\multicolumn{2}{c|}{}&
\multicolumn{2}{c|}{}&
\multicolumn{2}{c|}{}&
\multicolumn{2}{c|}{}&
\multicolumn{2}{c|}{}&
\multicolumn{2}{c|}{}&
\multicolumn{2}{c}{}\\
\noalign{\smallskip}
\hline
\noalign{\smallskip}
Initial Pose& \multicolumn{2}{c|}{-}& \multicolumn{2}{c|}{Ground Truth} &\multicolumn{2}{c|}{Ground Truth} &\multicolumn{2}{c|}{Ground Truth} &\multicolumn{2}{c|}{Ground Truth} &\multicolumn{2}{c|}{PoseCNN} &\multicolumn{2}{c}{Ground Truth}\\
Re-initialization& \multicolumn{2}{c|}{-}& \multicolumn{2}{c|}{No}& \multicolumn{2}{c|}{No}& \multicolumn{2}{c|}{Yes (290)}& \multicolumn{2}{c|}{No}& \multicolumn{2}{c|}{Yes (2)}&  \multicolumn{2}{c}{No}\\
\noalign{\smallskip}
\hline
\noalign{\smallskip}
Objects & ADD & ADD-S & ADD & ADD-S & ADD & ADD-S & ADD & ADD-S & ADD & ADD-S & ADD & ADD-S & ADD & ADD-S \\
\noalign{\smallskip}
\hline
\noalign{\smallskip}
002\_master\_chef\_can$^\star$  & 78.0 & \underline{96.3} & 55.6 & 90.7 & 46.2 & 90.2 & 89.0 & 93.8 & \textbf{93.9} & \underline{96.3} & \underline{89.3} & \textbf{96.7} & 66.4 & 89.7\\
003\_cracker\_box  & 91.4 & 95.3 & \underline{96.4} & \textbf{97.2} & 57.0 & 72.3 & 88.5 & 93.0 & \textbf{96.5} & \textbf{97.2} & 96.0 & 97.1 & 82.4 & 92.1\\
004\_sugar\_box  & \textbf{97.6} & \underline{98.2} & 97.1 & 97.9 & 50.4 & 72.7 & 94.3 & 96.3 & \textbf{97.6} & 98.1 & 94.0 & 96.4 & 96.1 & \textbf{98.4}\\
005\_tomato\_soup\_can$^\star$  & \underline{90.3} & 94.8 & 64.7 & 89.5 & 72.4 & 91.6 & 89.1 & 93.2 & \textbf{95.0} & \underline{97.2} & 87.2 & 95.2 & 73.2 & \textbf{97.3}\\
006\_mustard\_bottle  & \underline{97.1} & 98.0 & \underline{97.1} & 98.0 & 87.7 & 98.2 & 92.0 & 95.1 & 95.8 & 97.4 & \textbf{98.3} & \textbf{98.5} & 96.2 & \underline{98.4}\\
007\_tuna\_fish\_can$^\star$  & \textbf{92.2} & \textbf{98.0} & 69.1 & 93.3 & 28.7 & 52.9 & \underline{92.0} & \underline{96.4} & 86.5 & 91.1 & 86.8 & 93.6 & 73.2 & 95.8\\
008\_pudding\_box  & 83.5 & 90.6 & \underline{96.8} & \underline{97.9} & 12.7 & 18.0 & 80.1 & 88.3 & \textbf{97.9} & \textbf{98.4} & 60.9 & 87.1 & 73.8 & 88.9\\
009\_gelatin\_box  & \underline{98.0} & 98.5 & 97.5 & 98.4 & 49.1 & 70.7 & 92.0 & 94.4 & 97.8 & 98.4 & \textbf{98.2} & \underline{98.6} & 97.2 & \textbf{98.8}\\
010\_potted\_meat\_can  & 82.2 & \underline{90.3} & \underline{83.7} & 86.7 & 44.1 & 45.6 & 78.0 & 88.9 & 77.8 & 84.2 & 76.4 & 83.5 & \textbf{93.3} & \textbf{97.3}\\
011\_banana$^\diamond$  & \underline{94.9} & 97.6 & 86.3 & 96.1 & 93.3 & \underline{97.7} & 81.0 & 90.5 & \underline{94.9} & 97.2 & 92.8 & \underline{97.7} & \textbf{95.6} & \textbf{98.4}\\
019\_pitcher\_base$^\diamond$  & 97.4 & 97.9 & 97.3 & 97.7 & \textbf{97.9} & \underline{98.2} & 90.4 & 94.7 & 96.8 & 97.5 & \underline{97.7} & 98.1 & 97.0 & \textbf{98.8}\\
021\_bleach\_cleanser  & 91.6 & 96.9 & 95.2 & 97.2 & \textbf{95.9} & \underline{97.3} & 81.7 & 90.5 & \textbf{95.9} & 97.2 & \textbf{95.9} & 97.0 & 92.6 & \textbf{97.5}\\
024\_bowl$^\diamond$$^\star$  & 8.1 & 87.0 & 30.4 & \underline{97.2} & 24.2 & 82.4 & 38.8 & 90.6 & \textbf{80.9} & 94.5 & 34.0 & 93.0 & \underline{74.4} & \textbf{98.4}\\
025\_mug$^\diamond$  & \underline{94.2} & \underline{97.6} & 83.2 & 93.3 & 60.0 & 71.2 & 83.2 & 92.0 & 91.5 & 96.9 & 86.9 & 96.7 & \textbf{95.6} & \textbf{98.5}\\
035\_power\_drill  & 97.2 & 97.9 & 97.1 & 97.8 & \textbf{97.9} & \underline{98.3} & 85.4 & 92.3 & 96.4 & 97.4 & \underline{97.8} & 98.2 & 96.7 & \textbf{98.5}\\
036\_wood\_block  & 81.1 & 91.5 & \textbf{95.5} & \underline{96.9} & 45.7 & 62.5 & 44.3 & 75.4 & \underline{95.2} & 96.7 & 37.8 & 93.6 & 93.5 & \textbf{97.2}\\
037\_scissors$^\diamond$  & 92.7 & 96.0 & 4.2 & 16.2 & 20.9 & 38.6 & 70.3 & 84.5 & \textbf{95.7} & \textbf{97.5} & 72.7 & 85.5 & \underline{93.5} & \underline{97.3}\\
040\_large\_marker$^\star$  & 88.9 & \textbf{98.2} & 35.6 & 53.0 & 12.2 & 18.9 & 80.4 & 91.2 & \textbf{92.2} & 96.0 & \underline{89.2} & 97.3 & 88.5 & \underline{97.8}\\
051\_large\_clamp$^\diamond$  & 54.2 & 77.9 & 61.2 & 72.3 & 62.8 & 80.1 & 73.9 & 84.1 & \textbf{94.7} & \textbf{96.9} & 90.1 & 95.5 & \underline{91.8} & \textbf{96.9}\\
052\_extra\_large\_clamp$^\diamond$  & 36.5 & 77.8 & \textbf{93.7} & \textbf{96.6} & 67.5 & 69.7 & 49.3 & 90.3 & \underline{91.7} & \underline{95.8} & 84.4 & 94.1 & 85.9 & 94.3\\
061\_foam\_brick$^\diamond$  & 48.2 & 97.6 & \textbf{96.8} & 98.1 & 70.0 & 86.5 & 91.6 & 95.5 & 93.7 & 96.7 & 96.1 & \underline{98.3} & \underline{96.2} & \textbf{98.5}\\
\noalign{\smallskip}
\hline
\noalign{\smallskip}
\textbf{All Frames}  & 80.7 & 94.0 & 78.0 & 90.2 & 59.2 & 74.3 & 82.3 & 91.9 & \textbf{93.0} & \underline{95.7} & \underline{87.5} & 95.2 & 86.4 & \textbf{96.5}\\
\noalign{\smallskip}
\hline
\end{tabularx}
\end{table*}
For our algorithm, we use the parameters $\sigma_\textrm{r} = \{25,15,10\}$, $\sigma_\textrm{d} = \{50, 30, 20\}$, $s = \{7, 4, 2\}$, and $r_\textrm{t} = \{70, 50, 40\}$, where values are given in units of pixels and millimeters.
Our method is compared to \textit{PoseCNN} \cite{Xiang2018}, the particle-filter-based approaches of \cite{Wuethrich2013} and \cite{Issac2016}, and the current state of the art in deep learning-based 3D object tracking \cite{Li2018}, \cite{Wen2020}, and \cite{Deng2021}.
The evaluation shows that \textit{ICG} achieves state-of-the-art results with respect to the \mbox{\textit{ADD-S}} metric, outperforming all other algorithms.
For the \textit{ADD} score, textureless methods have a significant disadvantage since, for some objects, the geometry is not conclusive.
It is, for example, not possible to determine the rotation of a rotationally symmetric object without using texture.
However, even with that handicap, \textit{ICG} surpasses the texture-based approaches of \textit{PoseCNN} and \textit{DeepIM}.
Also, it comes very close to the results of \textit{PoseRBPF}.
In the end, only \mbox{\textit{se(3)-TrackNet}} is able to perform significantly better.

To utilize the full frequency of modern cameras, track multiple objects simultaneously, and save resources, efficiency is essential in real-world applications.
We thus report the speed and the required hardware for all algorithms in \cref{tab:e01}.
\begin{table}
	\caption{
		Average speed in frames per second and hardware requirements for the CPU and GPU.
		Except for \textit{PoseRBPF}\cite{Deng2021}, results are taken from \cite{Wen2020}.
		\textit{PoseRBPF} was evaluated without \textit{SDF}.
	}\label{tab:e01}

\centering
\small
\begin{tabularx}{\linewidth}{l@{\hspace{-0.2cm}} >{\centering\arraybackslash}X@{\hspace{-0.2cm}} >{\centering\arraybackslash}X@{\hspace{-0.25cm}} r}
		\hline
		\noalign{\smallskip}
		\textbf{Approach} & Single~Core & No GPU  & FPS\\
		\noalign{\smallskip}
		\hline
		\noalign{\smallskip}
		PoseCNN+ICP+DeepIM\cite{Xiang2018} &  & \xmark & 0.1 Hz\\
		Wüthrich\cite{Wuethrich2013} & \cmark & \cmark & 12.9 Hz\\
		RGF\cite{Issac2016} & \cmark & \cmark & 11.8 Hz\\
		DeepIM\cite{Li2018} &  & \xmark & 12.0 Hz\\
		se(3)-TrackNet\cite{Wen2020} &  & \xmark & 90.9 Hz\\
		PoseRBPF\cite{Deng2021} &  & \xmark & 7.6 Hz\\
		ICG (Ours) & \cmark & \cmark & 788.4 Hz\\
		\noalign{\smallskip}
		\hline
\end{tabularx}

\end{table}
The evaluation of \textit{ICG} is conducted on an \textit{Intel Xeon E5-1630 v4} CPU.
In comparison, \cite{Wen2020} used an \textit{Intel Xeon E5-1660 v3} CPU and a \textit{NVIDIA Tesla K40c} GPU.
The results show the outstanding efficiency of our approach.
While \textit{ICG} runs only on a single CPU core, it is almost one order of magnitude faster than the second-best algorithm \textit{se(3)-TrackNet}, which requires a high-performance GPU.

\subsection{OPT Dataset}\label{ssec:e1}
While the \textit{YCB-Video} dataset features challenging sequences in real-world environments and a large number of objects, ground truth has limited accuracy and with a large threshold $e_\textrm{t} = 0.1\,\unit{m}$, the dataset mostly evaluates robustness.
Also, images do not contain motion blur, favoring texture-based methods.
The \textit{OPT} dataset \cite{Wu2017} nicely complements those properties.
It includes $6$ objects and consists of $552$ real-world sequences with significant motion blur.
Ground truth is obtained using a calibration board.
For the evaluation, the area under curve of the \textit{ADD} metric is used with a threshold of $r_\textrm{t} = 0.2d$, where $d$ is the largest distance between model vertices.
The final value is scaled between $0$ and $20$.
Following \cite{Wu2017}, we refer to it as \textit{AUC} score.

Evaluation results are reported in \cref{tab:e10}.
\begin{table}
	\caption{
		Results on the \textit{OPT} dataset \cite{Wu2017}.
		The reported \textit{AUC} scores are scaled between zero and twenty to match other evaluations.
		Results are taken from \cite{Wu2017} and the respective publications.
	}\label{tab:e10}

\centering
\scriptsize
\begin{tabularx}{\linewidth}{@{\hspace{0.15cm}} l@{\hspace{-0.15cm}} *{5}{>{\centering\arraybackslash}X@{\hspace{-0.4cm}}} >{\centering\arraybackslash}X@{\hspace{-0.1cm}} c@{\hspace{0.15cm}}}
\hline
\noalign{\smallskip}
\textbf{Approach}&Soda & Chest & Ironman & House & Bike & Jet &\textbf{Avg.}\\
\noalign{\smallskip}
\hline
\noalign{\smallskip}
PWP3D\cite{Prisacariu2012}&5.87&5.55&3.92&3.58&5.36&5.81&5.01\\
ElasticFusion\cite{Whelan2015}&1.90&1.53&1.69&2.70&1.57&1.86&1.87\\
UDP\cite{Brachmann2016}&8.49&6.79&5.25&5.97&6.10&2.34&5.82\\
ORB-SLAM2\cite{MurArtal2017}&13.44&15.53&11.20&\underline{17.28}&10.41&9.93&12.97\\
Bugaev\cite{Bugaev2018}&14.85&14.97&14.71&14.48&12.55&\textbf{17.17}&14.79\\
Tjaden\cite{Tjaden2018}&8.86&11.76&11.99&10.15&11.90&13.22&11.31\\
Zhong\cite{Zhong2020}&9.01&12.24&11.21&13.61&12.83&15.44&12.39\\
Li\cite{Li2021}&9.00&14.92&13.44&13.60&12.85&10.64&12.41\\
SRT3D\cite{Stoiber2021}&\textbf{15.64}&\textbf{16.30}&\underline{17.41}&16.36&\underline{13.02}&15.64&\underline{15.73}\\
ICG (Ours)&\underline{15.32}&\underline{15.85}&\textbf{17.86}&\textbf{17.92}&\textbf{16.36}&\underline{15.90}&\textbf{16.54}\\
\noalign{\smallskip}
\hline
\end{tabularx}
\end{table}
For \textit{ICG}, the parameters $\sigma_\textrm{r} = \{15, 5, 1.5\}$, $\sigma_\textrm{d} = \{35, 35, 25\}$, $s = \{6, 4, 1\}$, and $r_\textrm{t} = \{50, 20, 10\}$ are used.
Also, like in \cite{Stoiber2021}, we constrain the rotationally symmetric soda object using $\lambda_\textrm{r} = 70000$.
In the evaluation, \textit{ICG} is compared to state-of-the-art classical methods that use different sources of information, including region, edge, texture, and depth.
Our approach performs best or second-best for all objects and improves on the state of the art by a significant margin.

\subsection{Choi Dataset}\label{ssec:e2}
Finally, we also want to evaluate the accuracy of our method.
For this, the simulated dataset of Choi \cite{Choi2013}, which features four sequences and perfect ground truth, is used.
To evaluate the accuracy, \ac{RMS} errors in the x, y, and z directions and in the roll, pitch, and yaw angles are calculated.
The rotational and translational mean over all four sequences is reported in \cref{tab:e20}.
\begin{table}
	\caption{
		Mean \ac{RMS} errors for translation and rotation parameters on the Choi dataset \cite{Choi2013}.
		Results are from the respective papers.
	}\label{tab:e20}

\centering
\scriptsize
\begin{tabularx}{\linewidth}{l@{\hspace{-0.05cm}} *{5}{>{\centering\arraybackslash}X@{\hspace{-0.1cm}}}}
\hline
\noalign{\smallskip}
\textbf{Approach} & Choi\cite{Choi2013} & Krull\cite{Krull2015} & Tan\cite{Tan2017} & Kehl\cite{Kehl2017} &ICG (Ours) \\
\noalign{\smallskip}
\hline
\noalign{\smallskip}
Translation [mm]& 1.36& 0.82& \underline{0.10}& 0.51& \textbf{0.04}\\
Rotation [degree]& 2.45& 1.38& \underline{0.07}& 0.26& \textbf{0.04}\\
\noalign{\smallskip}
\hline
\end{tabularx}
\end{table}
Detailed results are provided in \cref{sec:a1}.

For our algorithm, we use the parameters $\sigma_\textrm{r} = \{5\}$, $\sigma_\textrm{d} = \{10, 1\}$, $s = \{2, 1\}$, and $r_\textrm{t} = \{10\}$.
Note that since the dataset provides perfect depth and uncluttered color images, results have to be considered as an upper bound.
Nevertheless, the experiments demonstrate that, with good enough data, the method is highly accurate.

\subsection{Ablation Study}\label{ssec:e3}
To demonstrate the importance of the algorithm's components, we perform an ablation study on all three datasets.
Average results of this evaluation are provided in \cref{tab:e030}.
\begin{table}
	\caption{
		Ablation study on critical components of our algorithm.
	}\label{tab:e030}

\centering
\scriptsize
\begin{tabularx}{\linewidth}{l@{\hspace{0.05cm}} | *{2}{>{\centering\arraybackslash}X}| >{\centering\arraybackslash}X | *{2}{>{\centering\arraybackslash}X}}
\hline
\noalign{\smallskip}
\textbf{Dataset} & \multicolumn{2}{c|}{Choi\cite{Choi2013}} & \multicolumn{1}{c|}{OPT\cite{Wu2017}} & \multicolumn{2}{c}{YCB-Video\cite{Xiang2018}}\\
\noalign{\smallskip}
\hline
\noalign{\smallskip}
\textbf{Experiment} & Trans. & Rot. & AUC & ADD & \multicolumn{1}{c}{ADD-S}\\
\noalign{\smallskip}
\hline
\noalign{\smallskip}
Original&0.04&0.04&16.54& 86.4 & 96.5\\
W/o Region& 0.06 & 0.04 &8.94 & 66.1 & 84.1\\
W/o Depth& 41.65& 23.39& 15.88 & 26.6 & 42.8\\
W/o Regularization& 0.04 & 0.04 & 14.48 & 72.0 & 91.5 \\
W/o Occlusion Hand.& - & - & - & 77.6 & 91.9 \\
\noalign{\smallskip}
\hline
\end{tabularx}
\end{table}
The experiments show that while the effect of the depth and region modality differs between the datasets, each modality significantly contributes to the final result.
Also, we observe that while values on the Choi dataset stay the same without regularization, it is very important to the challenging sequences of the \textit{OPT} and \textit{YCB-Video} dataset.
For the case without occlusion handling, similar results are obtained.

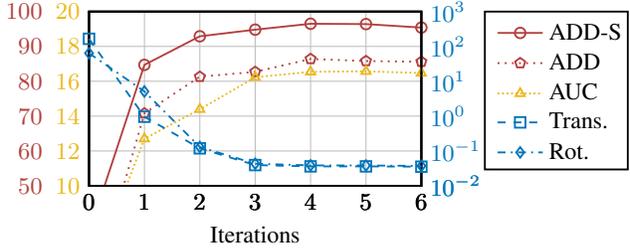
\begin{figure}[t]
	\centering

\centering
\small
\begin{tikzpicture}
\clip (-1.08,-0.8) rectangle + (8.25cm, 3.3);
\begin{axis}[
	legend style={at={(1.2,1.0)},anchor=north west, draw=none, legend cell align={left}},
	line width=0.7pt,
	width=6cm,
	height=3.9cm,
	xlabel near ticks,
	xlabel={Iterations},
	xmin=0, xmax=6,
	ymin=50, ymax=100,
	xtick={0,1,2,3,4,5,6},
	ytick={50,60,70,80,90,100},
	yticklabel style={dlrred, xshift=-0.5cm},
	ytick pos=left,
	xtick pos=bottom,
	tick style={draw=none},
	grid=both]
	\addplot[color=dlrred, every mark/.append style={solid}, mark=o] coordinates {
		(0,36.5)(1,84.7)(2,92.9)(3,94.8)(4,96.5)(5,96.4)(6,95.4)
	};
	\addlegendentry{ADD-S}
	\addplot[color=dlrred, dotted, every mark/.append style={solid}, mark=pentagon] coordinates {
		(0,15.5)(1,70.8)(2,81.3)(3,82.7)(4,86.4)(5,85.8)(6,85.6)
	};
	\addlegendentry{ADD}
\end{axis}	
\begin{axis}[
	legend style={at={(1.2,0.53)},anchor=west, draw=none, legend cell align={left}},
	line width=0.7pt,
	width=6cm,
	height=3.9cm,
	xmin=0, xmax=6,
	ymin=10, ymax=20,
	xtick={0,1,2,3,4,5,6},
	ytick={10,12,14,16,18,20},
	yticklabel style={dlrorange},
	ytick pos=left,
	xtick pos=bottom,
	tick style={draw=none}]
	\addplot[color=dlrorange, densely dotted, every mark/.append style={solid}, mark=triangle] coordinates {
		(0,4.42)(1,12.70)(2,14.39)(3,16.23)(4,16.54)(5,16.57)(6,16.47)
	};
	\addlegendentry{AUC}
\end{axis}
\begin{axis}[
	legend style={at={(1.2,0.07)},anchor=south west, draw=none, legend cell align={left}},
	ymode=log,
	line width=0.7pt,
	width=6cm,
	height=3.9cm,
	xmin=0, xmax=6,
	ymin=0.01, ymax=1000,
	xtick={0,1,2,3,4,5,6},
	ytick={0.01,0.01,0.1,1,10,100, 1000},
	yticklabel style={dlrblue},
	ytick pos=right,
	xtick pos=bottom,
	tick style={draw=none}]
	\addplot[color=dlrblue, dashed, every mark/.append style={solid}, mark = square] coordinates {
		(0,165.32)(1,0.96)(2,0.119)(3,0.0394)(4,0.0361)(5,0.0361)(6,0.0354)
	};
	\addlegendentry{Trans.}
	\addplot[color=dlrblue, dashdotted, every mark/.append style={solid}, mark = diamond] coordinates {
		(0,63.8)(1,5.09)(2,0.13)(3,0.043)(4,0.0385)(5,0.0383)(6,0.0372)
	};
	\addlegendentry{Rot.}
\end{axis}
\node[rectangle, draw=black, line width=0.7pt, minimum width = 1.9cm, minimum height = 2.1cm, anchor = north west] at (5.25,2.33) {};
\end{tikzpicture}
	\caption{
		Convergence plot showing the final results for the \textit{YCB-Video}, \textit{OPT}, and Choi dataset in red, yellow, and blue, respectively.
	} \label{fig:e01}
\end{figure}

Finally, the convergence of our approach is evaluated in \cref{fig:e01}.
For the \textit{YCB-Video} dataset, which values robustness, the algorithm converges after only two iterations.
With one additional iteration, accurate results are obtained on the Choi and \textit{OPT} datasets.
Independent of accuracy and robustness, fast convergence ensures that a maximum of only four iterations is sufficient to obtain excellent results.

\subsection{Limitations}\label{ssec:e4}
While \textit{ICG} achieves remarkable results, some limitations remain.
First, our method requires the geometry of tracked objects.
Also, for the region modality, objects have to be distinguishable from the background.
In addition, the depth camera has to provide reasonable measurements for the object's surface.
Another important limitation emerges if the object geometry is very similar in the vicinity of a particular pose.
Naturally, using geometric information alone, it is impossible to predict the correct pose in such cases.
Finally, like many conventional approaches that use line search, the algorithm has only a local view of the six-dimensional joint probability distribution.
As a consequence, it is constrained by local minima with a limit to the maximum pose difference between consecutive frames.

\section{Conclusion}\label{sec:c}

In this work, we developed \textit{ICG}, a highly efficient, textureless approach to 3D object tracking.
The method fuses region and depth in a well-founded probabilistic formulation that is able to handle occlusions in a robust manner.
While the overall algorithm is relatively simple and requires little computation, it performs surprisingly well on multiple datasets, outperforming the current state of the art for many cases in robustness, accuracy, and efficiency.
This is even more remarkable since \textit{ICG} has an inherent disadvantage in not using texture.
Assuming that texture would further improve results, our evaluation suggests that deep learning-based techniques do not yet surpass classical methods.
This is especially surprising since, at the expense of high computational cost and limited real-time capability, such algorithms can, in theory, directly consider all available information.
As a consequence, we believe that for both classical and learning-based tracking, as well as potential combinations, there remains a large number of ideas that wait to be explored to further improve 3D object tracking.

{\small
\bibliographystyle{ieee_fullname}
\bibliography{content/literature}
}

\vfill
\pagebreak
\appendix
\section*{Appendix}
In the following, we first state timings for the individual steps of our algorithm.
After this, the full results on the Choi dataset \cite{Choi2013} are presented, for which a concise version was shown in the main paper.
Subsequently, the developed region modality is evaluated on the \textit{RBOT} dataset \cite{Tjaden2018}, demonstrating improved tracking success.
Also, we compare to state-of-the-art \acs{6DoF} pose estimation algorithms on the \textit{YCB-Video} dataset \cite{Xiang2018} and discuss the role of 3D object tracking.
Finally, using predictions from modern pose estimation algorithms, we demonstrate that \textit{ICG} is well-suited for highly efficient pose refinement.

\section{Timings}\label{sec:a0}
In \cref{tab:e01}, an average framerate of $788.4\,\unit{Hz}$ was given for the evaluation on the \textit{YCB-Video} dataset \cite{Xiang2018}.
This corresponds to a total duration of $1.27\,\unit{ms}$ per frame.
Of this time, the algorithm needs $0.52\,\unit{ms}$ for the computation of correspondence lines, $0.58\,\unit{ms}$ for correspondence points, $0.09\,\unit{ms}$ for the calculation of gradient vectors and Hessian matrices, $0.05\,\unit{ms}$ for the update of color histograms, and the remaining $0.03\,\unit{ms}$ for other operations such as the optimization and pose update.
The timings demonstrate that the region- and depth-modality are well balanced, requiring similar amounts of computation.

\section{Choi Dataset}\label{sec:a1}
In the main paper in \cref{tab:e20}, only the averages over rotational and translational \acs{RMS} errors were presented for the Choi dataset \cite{Choi2013}.
For the sake of completeness, we also want to provide the full results with respect to the errors in the x, y, and z directions and in the roll, pitch, and yaw angles.
The results for each of the four evaluated objects as well as the mean values are shown in \cref{tab:a00}.
\begin{table}
	\caption{
		\acs{RMS} errors for translation and rotation parameters on the Choi dataset \cite{Choi2013}.
		Results are from the respective papers.
	}\label{tab:a00}

\centering
\scriptsize
\begin{tabularx}{\linewidth}{l l@{\hspace{-0.05cm}} *{5}{>{\centering\arraybackslash}X@{\hspace{-0.1cm}}}}
\hline
\noalign{\smallskip}
\multicolumn{2}{l}{\textbf{Approach}} & Choi\cite{Choi2013} & Krull\cite{Krull2015} & Tan\cite{Tan2017} & Kehl\cite{Kehl2017} &ICG (Ours) \\
\noalign{\smallskip}
\hline
\noalign{\smallskip}
\multirow{6}{0.8cm}{Kinect Box} &
X & 1.84 & 0.83 & \underline{0.15} & 0.76 & \textbf{0.05}\\
&Y & 2.23 & 1.67 & \underline{0.19} & 1.09 & \textbf{0.11}\\
&Z & 1.36 & 0.79 & \underline{0.09} & 0.38 & \textbf{0.03}\\
&Roll & 6.41 & 1.11 & \underline{0.09} & 0.17 & \textbf{0.02}\\
&Pitch & 0.76 & 0.55 & \underline{0.06} & 0.18 & \textbf{0.02}\\
&Yaw & 6.32 & 1.04 & \underline{0.04} & 0.20 & \textbf{0.02}\\
\noalign{\smallskip}
\hline
\noalign{\smallskip}
\multirow{6}{0.8cm}{Milk} &
X & 0.93 & 0.51 & \underline{0.09} & 0.64 & \textbf{0.02}\\
&Y & 1.94 & 1.27 & \underline{0.11} & 0.59 & \textbf{0.05}\\
&Z & 1.09 & 0.62 & \underline{0.08} & 0.24 & \textbf{0.02}\\
&Roll & 3.83 & 2.19 & \underline{0.07} & 0.41 & \textbf{0.06}\\
&Pitch & 1.41 & 1.44 & \underline{0.09} & 0.29 & \textbf{0.04}\\
&Yaw & 3.26 & 1.90 & \textbf{0.06} & 0.42 & \textbf{0.06}\\
\noalign{\smallskip}
\hline
\noalign{\smallskip}
\multirow{6}{0.8cm}{Orange Juice} &
X & 0.96 & 0.52 & \underline{0.11} & 0.50 & \textbf{0.04}\\
&Y & 1.44 & 0.74 & \underline{0.09} & 0.69 & \textbf{0.03}\\
&Z & 1.17 & 0.63 & \underline{0.09} & 0.17 & \textbf{0.02}\\
&Roll & 1.32 & 1.28 & \underline{0.08} & 0.12 & \textbf{0.05}\\
&Pitch & 0.75 & 1.08 & \underline{0.08} & 0.20 & \textbf{0.03}\\
&Yaw & 1.39 & 1.20 & \underline{0.08} & 0.19 & \textbf{0.06}\\
\noalign{\smallskip}
\hline
\noalign{\smallskip}
\multirow{6}{0.8cm}{Tide} &
X & 0.83 & 0.69 & \underline{0.08} & 0.34 & \textbf{0.02}\\
&Y & 1.37 & 0.81 & \underline{0.09} & 0.49 & \textbf{0.03}\\
&Z & 1.20 & 0.81 & \underline{0.07} & 0.18 & \textbf{0.01}\\
&Roll & 1.78 & 2.10 & \underline{0.05} & 0.15 & \textbf{0.03}\\
&Pitch & 1.09 & 1.38 & \underline{0.12} & 0.39 & \textbf{0.04}\\
&Yaw & 1.13 & 1.27 & \underline{0.05} & 0.37 & \textbf{0.03}\\
\noalign{\smallskip}
\hline
\noalign{\smallskip}
\multicolumn{2}{l}{Mean Translation}& 1.36& 0.82& \underline{0.10}& 0.51& \textbf{0.04}\\
\multicolumn{2}{l}{Mean Rotation}& 2.45& 1.38& \underline{0.07}& 0.26& \textbf{0.04}\\
\noalign{\smallskip}
\hline
\end{tabularx}
\end{table}

\section{RBOT Dataset}\label{sec:a2}
In \cref{ssec:p2}, we modified the region-based approach of \textit{SRT3D} \cite{Stoiber2020b, Stoiber2021} to be independent of the scale space and to incorporate a user-defined uncertainty.
In the following, we want to show that this is not only convenient for the combination with the depth modality but that the modifications also improve tracking results.
We thereby use the \textit{RBOT} dataset \cite{Tjaden2018} to compare our approach to the state of the art in region-based tracking as well as to additional methods that include edge information.
\begin{table*}
	\caption{
		Tracking success rate for all objects and scenarios on the \textit{RBOT} dataset \cite{Tjaden2018}.
		Methods that incorporate information from edges in addition to region are indicated by a $^\star$.
		Results are from the respective publications.
	}\label{tab:a01}

\centering
\scriptsize
\begin{tabularx}{\textwidth}{@{\hspace{0.15cm}} l@{\hspace{-0.1cm}} *{17}{>{\centering\arraybackslash}X@{\hspace{-0.4cm}}} >{\centering\arraybackslash}X@{\hspace{-0.0cm}} c@{\hspace{0.15cm}}}
\hline
\noalign{\smallskip}
\textbf{Approach}& Ape & Soda & Vise & Soup & Camera & Can & Cat & Clown & Cube & Driller & Duck & Egg Box & Glue & Iron & Candy & Lamp & Phone & Squirrel &\textbf{Avg.}\\
\noalign{\smallskip}
\hline
\noalign{\medskip}

\noalign{Regular}
\noalign{\medskip}
Tjaden \cite{Tjaden2018} & 85.0& 39.0& 98.9& 82.4& 79.7& 87.6& 95.9& 93.3& 78.1& 93.0& 86.8& 74.6& 38.9& 81.0& 46.8 & \underline{97.5}& 80.7& 99.4& 79.9\\
Zhong \cite{Zhong2020} & 88.8& 41.3& 94.0& 85.9& 86.9& 89.0& 98.5& 93.7& 83.1& 87.3& 86.2& 78.5& 58.6& 86.3& 57.9& 91.7& 85.0& 96.2& 82.7\\
Huang \cite{Huang2020}$^\star$ & 91.9& 44.8 & \textbf{99.7}& 89.1& 89.3& 90.6& 97.4& 95.9& 83.9 & \underline{97.6}& 91.8& 84.4& 59.0& 92.5& 74.3& 97.4& 86.4& 99.7& 86.9\\
Liu \cite{Liu2021}$^\star$ & 93.7& 39.3& 98.4& 91.6& 84.6& 89.2& 97.9& 95.9& 86.3& 95.1& 93.4& 77.7& 61.5& 87.8& 65.0& 95.2& 85.7 & \underline{99.8}& 85.5\\
Li \cite{Li2021}$^\star$ & 92.8& 42.6& 96.8& 87.5& 90.7& 86.2& 99.0& 96.9& 86.8& 94.6& 90.4& 87.0& 57.6& 88.7& 59.9& 96.5& 90.6& 99.5& 85.8\\
Sun \cite{Sun2021}$^\star$ & 93.0& 55.2& 99.3& 85.4& 96.1& 93.9& 98.0& 95.6& 79.5 & \textbf{98.2}& 89.7& 89.1& 66.5& 91.3& 60.6 & \textbf{98.6}& 95.6& 99.6& 88.1\\
SRT3D \cite{Stoiber2021}  & \textbf{98.8} & \underline{65.1} & \underline{99.6} & \textbf{96.0} & \textbf{98.0} & \underline{96.5} & \textbf{100.0} & \underline{98.4} & \underline{94.1}& 96.9 & \textbf{98.0} & \underline{95.3} & \underline{79.3} & \textbf{96.0} & \underline{90.3}& 97.4 & \underline{96.2} & \underline{99.8} & \underline{94.2}\\
ICG (Ours)  & \underline{98.1} & \textbf{66.4} & \underline{99.6} & \textbf{96.0} & \underline{97.4} & \textbf{96.9} & \textbf{100.0} & \textbf{98.5} & \textbf{94.8} & \underline{97.6} & \textbf{98.0} & \textbf{95.5} & \textbf{80.8} & \underline{95.9} & \textbf{91.0}& 97.1 & \textbf{96.6} & \textbf{99.9} & \textbf{94.4}\\
\hline
\noalign{\medskip}

\noalign{Dynamic Light}
\noalign{\medskip}
Tjaden \cite{Tjaden2018} & 84.9& 42.0& 99.0& 81.3& 84.3& 88.9& 95.6& 92.5& 77.5& 94.6& 86.4& 77.3& 52.9& 77.9& 47.9& 96.9& 81.7& 99.3& 81.2\\
Zhong \cite{Zhong2020} & 89.7& 40.2& 92.7& 86.5& 86.6& 89.2& 98.3& 93.9& 81.8& 88.4& 83.9& 76.8& 55.3& 79.3& 54.7& 88.7& 81.0& 95.8& 81.3\\
Huang \cite{Huang2020}$^\star$ & 91.8& 42.3& 98.9& 89.9& 91.3& 87.8& 97.6& 94.5& 84.5 & \underline{98.1}& 91.9& 86.7& 66.2& 90.9& 73.2& 97.1& 89.2& 99.6& 87.3\\
Liu \cite{Liu2021}$^\star$ & 93.5& 38.2& 98.4& 88.8& 87.0& 88.5& 98.1& 94.4& 85.1& 95.1& 92.7& 76.1& 58.1& 79.6& 62.1& 93.2& 84.7& 99.6& 84.1\\
Li \cite{Li2021}$^\star$ & 93.5& 43.1& 96.6& 88.5& 92.8& 86.0& 99.6& 95.5& 85.7& 96.8& 91.1& 90.2& 68.4& 86.8& 59.7& 96.1& 91.5& 99.2& 86.7\\
Sun \cite{Sun2021}$^\star$ & 93.8& 55.9 & \textbf{99.6}& 85.6 & \textbf{97.7}& 93.7& 97.7& 96.5& 78.3 & \textbf{98.6}& 91.0& 91.6& 72.1& 90.7& 63.0 & \textbf{98.9}& 94.4 & \textbf{100.0}& 88.8\\
SRT3D \cite{Stoiber2021}  & \underline{98.2} & \underline{65.2}& 99.2 & \underline{95.6}& 97.5 & \textbf{98.1} & \textbf{100.0} & \underline{98.5} & \underline{94.2}& 97.5 & \textbf{97.9} & \underline{96.9} & \textbf{86.1} & \underline{95.2} & \underline{89.3}& 97.0 & \underline{95.9} & \underline{99.9} & \underline{94.6}\\
ICG (Ours)  & \textbf{98.4} & \textbf{67.0} & \underline{99.5} & \textbf{95.7} & \underline{97.6} & \underline{97.5} & \underline{99.8} & \textbf{98.6} & \textbf{94.9}& 97.5 & \underline{97.4} & \textbf{97.1} & \underline{85.5} & \textbf{96.0} & \textbf{91.5} & \underline{97.7} & \textbf{96.2} & \underline{99.9} & \textbf{94.9}\\
\hline
\noalign{\medskip}

\noalign{Noise}
\noalign{\medskip}
Tjaden \cite{Tjaden2018} & 77.5& 44.5& 91.5& 82.9& 51.7& 38.4& 95.1& 69.2& 24.4& 64.3& 88.5& 11.2& 2.9& 46.7& 32.7& 57.3& 44.1& 96.6& 56.6\\
Zhong \cite{Zhong2020} & 79.3& 35.2& 82.6& 86.2& 65.1& 56.9& 96.9& 67.0& 37.5& 75.2& 85.4& 35.2& 18.9& 63.7& 35.4& 64.6& 66.3& 93.2& 63.6\\
Huang \cite{Huang2020}$^\star$ & 89.0& 45.0& 89.5& 90.2& 68.9& 38.3& 95.9& 72.8& 20.1& 85.5& 92.2& 26.8& 15.8& 66.2& 52.2& 58.3& 65.1& 98.4& 65.0\\
Liu \cite{Liu2021}$^\star$ & 84.7& 33.0& 88.8& 89.5& 56.4& 50.1& 94.1& 66.5& 32.3& 79.6& 94.2& 29.6& 19.9& 63.4& 40.3& 61.6& 62.4& 96.9& 63.5\\
Li \cite{Li2021}$^\star$ & 89.1& 44.0& 91.6& 89.4& 75.2& 62.3& 98.6& 77.3& 41.2& 81.5& 91.6& 54.5& 31.8& 65.0& 46.0 & \underline{78.5}& 69.6& 97.6& 71.4\\
Sun \cite{Sun2021}$^\star$ & 92.5& 56.2 & \textbf{98.0}& 85.1 & \textbf{91.7} & \textbf{79.0}& 97.7& 86.2& 40.1 & \textbf{96.6}& 90.8 & \textbf{70.2} & \textbf{50.9} & \textbf{84.3}& 49.9 & \textbf{91.2} & \textbf{89.4} & \underline{99.4}& 80.5\\
SRT3D \cite{Stoiber2021}  & \underline{96.9} & \underline{61.9} & \underline{95.4} & \underline{95.7}& 84.5& 73.9 & \textbf{99.9} & \underline{90.3} & \textbf{62.2}& 87.8 & \textbf{97.6}& 62.2& 43.4 & \textbf{84.3} & \underline{78.2}& 73.3& 83.1 & \textbf{99.7} & \underline{81.7}\\
ICG (Ours)  & \textbf{98.0} & \textbf{64.3} & \underline{95.4} & \textbf{95.8} & \underline{84.8} & \underline{74.8} & \textbf{99.9} & \textbf{90.5} & \underline{61.9} & \underline{88.5} & \underline{97.4} & \underline{63.4} & \underline{45.3}& 84.2 & \textbf{81.2}& 74.0 & \underline{84.8} & \underline{99.4} & \textbf{82.4}\\
\hline
\noalign{\medskip}

\noalign{Unmodeled Occlusion}
\noalign{\medskip}
Tjaden \cite{Tjaden2018} & 80.0& 42.7& 91.8& 73.5& 76.1& 81.7& 89.8& 82.6& 68.7& 86.7& 80.5& 67.0& 46.6& 64.0& 43.6& 88.8& 68.6& 86.2& 73.3\\
Zhong \cite{Zhong2020} & 83.9& 38.1& 92.4& 81.5& 81.3& 85.5& 97.5& 88.9& 76.1& 87.5& 81.7& 72.7& 52.5& 77.2& 53.9& 88.5& 79.3& 92.5& 78.4\\
Huang \cite{Huang2020}$^\star$ & 86.2& 46.3& 97.8& 87.5& 86.5& 86.3& 95.7& 90.7& 78.8& 96.5& 86.0& 80.6& 59.9& 86.8& 69.6& 93.3& 81.8& 95.8& 83.6\\
Liu \cite{Liu2021}$^\star$ & 87.1& 36.7& 91.7& 78.8& 79.2& 82.5& 92.8& 86.1& 78.0& 90.2& 83.4& 72.0& 52.3& 72.8& 55.9& 86.9& 77.8& 93.0& 77.6\\
Li \cite{Li2021}$^\star$ & 89.3& 43.3& 92.2& 83.1& 84.1& 79.0& 94.5& 88.6& 76.2& 90.4& 87.0& 80.7& 61.6& 75.3& 53.1& 91.1& 81.9& 93.4& 80.3\\
Sun \cite{Sun2021}$^\star$ & 91.3& 56.7& 97.8& 82.0& 92.8& 89.9& 96.6& 92.2& 71.8 & \textbf{97.0}& 85.0& 84.6& 66.9& 87.7& 56.1& 95.1& 89.8& 98.2& 85.1\\
SRT3D \cite{Stoiber2021}  & \underline{96.5} & \textbf{66.8} & \underline{99.0} & \underline{95.8} & \textbf{95.0} & \underline{95.9} & \textbf{100.0} & \underline{97.6} & \underline{92.2} & \underline{96.6} & \underline{95.0} & \underline{94.4} & \underline{79.0} & \underline{94.7} & \textbf{89.8} & \underline{95.7} & \underline{93.6} & \textbf{99.6} & \underline{93.2}\\
ICG (Ours)  & \textbf{97.3} & \underline{66.3} & \textbf{99.3} & \textbf{96.0} & \textbf{95.0} & \textbf{96.5} & \textbf{100.0} & \textbf{97.7} & \textbf{92.9}& 96.4 & \textbf{96.1} & \textbf{96.5} & \textbf{82.1} & \textbf{96.1} & \underline{89.7} & \textbf{95.8} & \textbf{94.2} & \underline{99.2} & \textbf{93.7}\\
\hline
\noalign{\medskip}

\noalign{Modeled Occlusion}
\noalign{\medskip}
Tjaden \cite{Tjaden2018} & 82.0& 42.0& 95.7& 81.1& 78.7& 83.4& 92.8& 87.9& 74.3& 91.7& 84.8& 71.0& 49.1& 73.0& 46.3& 90.9& 76.2& 96.9& 77.7\\
Huang \cite{Huang2020}$^\star$ & 87.8& 45.5& 98.1& 87.2& 89.0& 89.8& 95.1& 91.4& 77.4 & \textbf{97.1}& 87.7& 83.0& 62.5& 88.6& 69.7& 94.1& 86.0& 98.9& 84.9\\
SRT3D \cite{Stoiber2021}  & \textbf{97.9} & \underline{68.3} & \underline{99.2} & \underline{95.4} & \underline{96.8} & \underline{96.4} & \underline{99.6} & \underline{98.6} & \underline{93.0}& 96.4 & \underline{96.6} & \underline{96.2} & \underline{82.9} & \underline{95.1} & \underline{91.0} & \underline{96.0} & \underline{94.5} & \underline{99.6} & \underline{94.1}\\
ICG (Ours)  & \textbf{97.9} & \textbf{69.1} & \textbf{99.5} & \textbf{97.2} & \textbf{97.1} & \textbf{96.9} & \textbf{99.9} & \textbf{98.9} & \textbf{93.2} & \underline{97.0} & \textbf{97.8} & \textbf{97.2} & \textbf{84.3} & \textbf{96.0} & \textbf{92.6} & \textbf{97.4} & \textbf{95.3} & \textbf{99.8} & \textbf{94.8}\\

\hline
\end{tabularx}

\end{table*}

All experiments in the evaluation are performed as defined by \cite{Tjaden2018}.
The required translational and rotational errors are calculated as follows
\begin{equation}\label{eq:a00}
	e_{\textrm{t}} = \big\lVert{}_\textrm{M}\pmb{t}_{\textrm{M}_\textrm{gt}} \big\rVert_2,
\end{equation}
\begin{equation}\label{eq:a01}
	e_{\textrm{r}} = \cos^{-1}\hspace{-2pt}\bigg(\frac{\trace( {}_\textrm{M}\pmb{R}_{\textrm{M}_\textrm{gt}}) - 1}{2}\bigg).
\end{equation}
Based on those errors, the tracking success is calculated as the percentage of estimated poses with 
$e_{\textrm{t}} < 5\,\unit{cm}$ and  \mbox{$e_{\textrm{r}} < 5^\circ$}.
In cases of unsuccessful tracking, the algorithm is re-initialized with the ground-truth pose.
For \textit{ICG}, we use the same parameter values as in \cite{Stoiber2021} and define a decreasing standard deviation of $\sigma_\textrm{r} = \{15,5,3.5,1.5\}$.

Results of the evaluation are shown in \cref{tab:a01}.
The reported scores show that both \textit{SRT3D} and \textit{ICG} achieve significantly higher results than the remaining algorithms.
However, on average, \textit{ICG} performs about half a percentage point better than \textit{SRT3D}.
This demonstrates that setting a defined standard deviation $\sigma_\textrm{r}$ instead of using an implicit variance of $\sigma^2 = \nicefrac{s_\textrm{h} s^2}{\bar{n}^2}$ helps to further improve results.

\section{6DoF Pose Estimation}\label{sec:a3}
Given the strong results of modern \acs{6DoF} pose estimation methods \cite{Labbe2020, He2021}, the question arises whether 3D object tracking is even necessary.
In order to answer this, we compare \textit{ICG} with state-of-the-art pose estimation methods on the \textit{YCB-Video} dataset \cite{Xiang2018}.
The \textit{ADD(S)} metric is thereby adopted to ensure compatibility with reported results from \textit{PVN3D} \cite{He2020} and \textit{FFB6D} \cite{He2021}.
It is a combined metric that uses the \textit{ADD-S} score for symmetric objects and the \textit{ADD} error in all other cases.

Results of the evaluation are shown in \cref{tab:a02}.
\begin{table*}
	\caption{
		Comparison against state-of-the-art \acs{6DoF} pose estimation methods on the \textit{YCB-Video} dataset \cite{Xiang2018} with \textit{ADD(S)} and \textit{ADD-S} area under curve scores in percent.
		Results for \textit{Augmented Autoencoders}\protect\footnotemark[2] \cite{Sundermeyer2018}, \textit{CosyPose}\protect\footnotemark[3] \cite{Labbe2020}, and \textit{ICG} were computed by us.
		All other values are from the respective publications.
		While \textit{CosyPose} was trained on real data, good results can also be obtained using synthetic data alone \cite{Hodan2020}.
		Symmetric objects for which the \textit{ADD(S)} metric reports the \textit{ADD-S} instead of the \textit{ADD} error are indicated by a $^\star$.
	}\label{tab:a02}

\centering
\scriptsize
\begin{tabularx}{\textwidth}{X | >{\centering\arraybackslash}p{0.81cm}@{\hspace{0.0cm}} >{\centering\arraybackslash}p{0.81cm} | >{\centering\arraybackslash}p{0.97cm}@{\hspace{0.0cm}} >{\centering\arraybackslash}p{0.97cm} *{5}{| >{\centering\arraybackslash}p{0.81cm}@{\hspace{0.0cm}} >{\centering\arraybackslash}p{0.81cm}}}
\hline
\noalign{\smallskip}
\multirow{2}{1cm}{\textbf{Approach}} & 
\multicolumn{2}{c|}{\multirow{2}{1.62cm}{\centering PoseCNN \cite{Xiang2018}}}& 
\multicolumn{2}{c|}{\multirow{2}{1.94cm}{\centering Augmented Autoencoders\protect\footnotemark[2] \cite{Sundermeyer2018}}}&
\multicolumn{2}{c|}{\multirow{2}{1.62cm}{\centering DenseFusion \cite{Wang2019b}}}&
\multicolumn{2}{c|}{\multirow{2}{1.62cm}{\centering CosyPose\protect\footnotemark[3] \cite{Labbe2020}}}&
\multicolumn{2}{c|}{\multirow{2}{1.62cm}{\centering PVN3D \cite{He2020}}}&
\multicolumn{2}{c|}{\multirow{2}{1.62cm}{\centering FFB6D \cite{He2021}}}&
\multicolumn{2}{c}{\multirow{2}{1.62cm}{\centering ICG (Ours)}}\\
&\multicolumn{2}{c|}{}&
\multicolumn{2}{c|}{}&
\multicolumn{2}{c|}{}&
\multicolumn{2}{c|}{}&
\multicolumn{2}{c|}{}&
\multicolumn{2}{c}{}\\
\noalign{\smallskip}
\hline
\noalign{\smallskip}
(Training) Data & \multicolumn{2}{c|}{Real RGB} & \multicolumn{2}{c|}{3D Model} & \multicolumn{2}{c|}{Real RGB-D} & \multicolumn{2}{c|}{Real RGB} & \multicolumn{2}{c|}{Real RGB-D} & \multicolumn{2}{c|}{Real RGB-D} & \multicolumn{2}{c}{3D Model}\\
\noalign{\smallskip}
\hline
\noalign{\smallskip}
Objects & \multicolumn{2}{c|}{ADD(S) ADD-S} & \multicolumn{2}{c|}{ADD(S) ADD-S} &\multicolumn{2}{c|}{ADD(S) ADD-S} &\multicolumn{2}{c|}{ADD(S) ADD-S} &\multicolumn{2}{c|}{ADD(S) ADD-S} &\multicolumn{2}{c|}{ADD(S) ADD-S} &\multicolumn{2}{c}{ADD(S) ADD-S} \\
\noalign{\smallskip}
\hline
\noalign{\smallskip}
002\_master\_chef\_can  & 50.9 & 84.0 & 27.1 & 50.6 & - & \textbf{96.4} & 37.3 & 90.6 & \underline{80.5} & 96.0 & \textbf{80.6} & \underline{96.3} & 66.4 & 89.7\\
003\_cracker\_box  & 51.7 & 76.9 & 32.2 & 64.5 & - & 95.5 & 76.8 & 94.9 & \textbf{94.8} & \underline{96.1} & \underline{94.6} & \textbf{96.3} & 82.4 & 92.1\\
004\_sugar\_box  & 68.6 & 84.3 & 73.6 & 88.6 & - & 97.5 & 95.2 & \underline{97.6} & \underline{96.3} & 97.4 & \textbf{96.6} & \underline{97.6} & 96.1 & \textbf{98.4}\\
005\_tomato\_soup\_can  & 66.0 & 80.9 & 72.3 & 84.4 & - & 94.6 & \textbf{90.5} & 94.6 & 88.5 & \underline{96.2} & \underline{89.6} & 95.6 & 73.2 & \textbf{97.3}\\
006\_mustard\_bottle  & 79.9 & 90.2 & 77.5 & 90.9 & - & 97.2 & 92.7 & 96.5 & \underline{96.2} & 97.5 & \textbf{97.0} & \underline{97.8} & \underline{96.2} & \textbf{98.4}\\
007\_tuna\_fish\_can  & 70.4 & 87.9 & 71.2 & 92.2 & - & 96.6 & \textbf{93.9} & \textbf{97.5} & \underline{89.3} & 96.0 & 88.9 & \underline{96.8} & 73.2 & 95.8\\
008\_pudding\_box  & 62.9 & 79.0 & 47.9 & 67.7 & - & 96.5 & 93.5 & 96.2 & \textbf{95.7} & \textbf{97.1} & \underline{94.6} & \textbf{97.1} & 73.8 & 88.9\\
009\_gelatin\_box  & 75.2 & 87.1 & 74.8 & 82.9 & - & \underline{98.1} & 94.1 & 96.1 & 96.1 & 97.7 & \underline{96.9} & \underline{98.1} & \textbf{97.2} & \textbf{98.8}\\
010\_potted\_meat\_can  & 59.6 & 78.5 & 53.6 & 63.3 & - & 91.3 & 75.9 & 84.0 & \underline{88.6} & 93.3 & 88.1 & \underline{94.7} & \textbf{93.3} & \textbf{97.3}\\
011\_banana  & 72.3 & 85.9 & 13.1 & 51.6 & - & 96.6 & 90.0 & 95.6 & 93.7 & 96.6 & \underline{94.9} & \underline{97.2} & \textbf{95.6} & \textbf{98.4}\\
019\_pitcher\_base  & 52.5 & 76.8 & 77.6 & 91.7 & - & 97.1 & 94.0 & 97.3 & 96.5 & 97.4 & \underline{96.9} & \underline{97.6} & \textbf{97.0} & \textbf{98.8}\\
021\_bleach\_cleanser  & 50.5 & 71.9 & 42.0 & 62.6 & - & 95.8 & 82.1 & 92.7 & \underline{93.2} & 96.0 & \textbf{94.8} & \underline{96.8} & 92.6 & \textbf{97.5}\\
024\_bowl$^\star$  & 69.7 & 69.7 & 79.1 & 79.1 & - & 88.2 & 87.8 & 87.8 & 90.2 & 90.2 & \underline{96.3} & \underline{96.3} & \textbf{98.4} & \textbf{98.4}\\
025\_mug  & 57.7 & 78.0 & 58.0 & 80.9 & - & 97.1 & 87.8 & 94.9 & \underline{95.4} & \underline{97.6} & 94.2 & 97.3 & \textbf{95.6} & \textbf{98.5}\\
035\_power\_drill  & 55.1 & 72.8 & 61.2 & 77.9 & - & 96.0 & 89.7 & 95.1 & 95.1 & 96.7 & \underline{95.9} & \underline{97.2} & \textbf{96.7} & \textbf{98.5}\\
036\_wood\_block$^\star$  & 65.8 & 65.8 & 55.2 & 55.2 & - & 89.7 & 80.5 & 80.5 & 90.4 & 90.4 & \underline{92.6} & \underline{92.6} & \textbf{97.2} & \textbf{97.2}\\
037\_scissors  & 35.8 & 56.2 & 0.8 & 7.0 & - & 95.2 & 67.6 & 81.5 & 92.7 & 96.7 & \textbf{95.7} & \textbf{97.7} & \underline{93.5} & \underline{97.3}\\
040\_large\_marker  & 58.0 & 71.4 & 55.6 & 67.6 & - & \underline{97.5} & 84.3 & 93.1 & \textbf{91.8} & 96.7 & \underline{89.1} & 96.6 & 88.5 & \textbf{97.8}\\
051\_large\_clamp$^\star$  & 49.9 & 49.9 & 72.2 & 72.2 & - & 72.9 & 91.3 & 91.3 & 93.6 & 93.6 & \underline{96.8} & \underline{96.8} & \textbf{96.9} & \textbf{96.9}\\
052\_extra\_large\_clamp$^\star$  & 47.0 & 47.0 & 59.5 & 59.5 & - & 69.8 & 75.7 & 75.7 & 88.4 & 88.4 & \textbf{96.0} & \textbf{96.0} & \underline{94.3} & \underline{94.3}\\
061\_foam\_brick$^\star$  & 87.8 & 87.8 & 56.2 & 56.2 & - & 92.5 & 94.7 & 94.7 & 96.8 & 96.8 & \underline{97.3} & \underline{97.3} & \textbf{98.5} & \textbf{98.5}\\
\noalign{\smallskip}
\hline
\noalign{\smallskip}
\textbf{All Frames}  & 60.0 & 75.9 & 57.5 & 72.8 & - & 93.1 & 83.8 & 92.6 & \underline{91.8} & 95.5 & \textbf{92.7} & \textbf{96.6} & 87.9 & \underline{96.5}\\
\noalign{\smallskip}
\hline
\end{tabularx}
\end{table*}%
\begin{table*}
	\caption{
		Refined and unrefined results on the \textit{YCB-Video} dataset \cite{Xiang2018} with \textit{ADD} and \textit{ADD-S} area under curve scores in percent.
		For \textit{PoseCNN} with multi-hypothesis \textit{ICP}, results are taken from the corresponding publication \cite{Xiang2018}.
		To evaluate the refinement, predicted poses for \textit{PoseCNN} \cite{Xiang2018} are taken from the \textit{YCB\_Video\_toolbox}\protect\footnotemark[1] while results for \textit{Augmented Autoencoders}\protect\footnotemark[2] \cite{Sundermeyer2018} and \textit{CosyPose}\protect\footnotemark[3] \cite{Labbe2020} are computed using source code from the respective repositories.
	}\label{tab:a03}

\centering
\scriptsize
\begin{tabularx}{\textwidth}{X *{7}{| >{\centering\arraybackslash}p{0.83cm}@{\hspace{0.0cm}} >{\centering\arraybackslash}p{0.83cm}}}
\hline
\noalign{\smallskip}
\textbf{Approach} & 
\multicolumn{2}{c|}{\centering PoseCNN \cite{Xiang2018}}&
\multicolumn{4}{c|}{\centering PoseCNN\protect\footnotemark[1] \cite{Xiang2018}}&
\multicolumn{4}{c|}{\centering Augmented Autoencoders\protect\footnotemark[2] \cite{Sundermeyer2018}}&
\multicolumn{4}{c}{\centering CosyPose\protect\footnotemark[3] \cite{Labbe2020}}\\
\noalign{\smallskip}
\hline
\noalign{\smallskip}
\textbf{Refinement} & 
\multicolumn{2}{c|}{\centering MH ICP}&
\multicolumn{2}{c|}{\centering-}&
\multicolumn{2}{c|}{\centering ICG (Ours)}&
\multicolumn{2}{c|}{\centering-}&
\multicolumn{2}{c|}{\centering ICG (Ours)}&
\multicolumn{2}{c|}{\centering Iterative Matching}&
\multicolumn{2}{c}{\centering IM + ICG (Ours)}\\
\noalign{\smallskip}
\hline
\noalign{\smallskip}
Objects & ADD & ADD-S & ADD & ADD-S & ADD & ADD-S & ADD & ADD-S & ADD & ADD-S & ADD & ADD-S & ADD & ADD-S \\
\noalign{\smallskip}
\hline
\noalign{\smallskip}
002\_master\_chef\_can  & \textbf{69.0} & \textbf{95.8} & 50.0 & 84.6 & \underline{66.7} & \underline{94.7} & 27.1 & 50.6 & 38.3 & 67.2 & 37.3 & 90.6 & 38.6 & 93.1\\
003\_cracker\_box  & \underline{80.7} & 91.8 & 53.0 & 77.5 & 67.8 & 85.6 & 32.2 & 64.5 & 43.8 & 71.6 & 76.8 & \underline{94.9} & \textbf{81.4} & \textbf{97.9}\\
004\_sugar\_box  & \textbf{97.2} & \textbf{98.2} & 68.3 & 84.5 & 91.9 & 96.3 & 73.6 & 88.6 & 85.2 & 94.9 & 95.2 & 97.6 & \underline{95.8} & \textbf{98.2}\\
005\_tomato\_soup\_can  & 81.6 & 94.5 & 66.1 & 81.4 & 82.8 & 91.3 & 72.3 & 84.4 & 82.3 & 90.0 & \underline{90.5} & \underline{94.6} & \textbf{92.6} & \textbf{95.9}\\
006\_mustard\_bottle  & \textbf{97.0} & \textbf{98.4} & 80.8 & 91.1 & 93.9 & 97.4 & 77.5 & 90.9 & 87.9 & 96.9 & 92.7 & 96.5 & \underline{96.3} & \textbf{98.4}\\
007\_tuna\_fish\_can  & 83.1 & \underline{97.1} & 70.5 & 88.4 & 82.2 & 93.5 & 71.2 & 92.2 & 78.6 & 95.2 & \textbf{93.9} & \textbf{97.5} & \underline{92.2} & 95.6\\
008\_pudding\_box  & \textbf{96.6} & \textbf{97.9} & 62.2 & 79.3 & 72.3 & 85.1 & 47.9 & 67.7 & 58.6 & 81.7 & \underline{93.5} & \underline{96.2} & 81.9 & 91.6\\
009\_gelatin\_box  & \textbf{98.2} & \textbf{98.8} & 74.9 & 87.7 & \underline{95.1} & 97.8 & 74.8 & 82.9 & 81.9 & 88.9 & 94.1 & 96.1 & 93.5 & \underline{97.9}\\
010\_potted\_meat\_can  & \textbf{83.8} & \textbf{92.8} & 59.3 & 78.8 & 69.1 & 82.4 & 53.6 & 63.3 & 61.7 & 68.0 & 75.9 & 84.0 & \underline{78.8} & \underline{85.7}\\
011\_banana  & \underline{91.6} & \underline{96.9} & 72.3 & 86.3 & 80.4 & 92.0 & 13.1 & 51.6 & 18.2 & 60.1 & 90.0 & 95.6 & \textbf{95.2} & \textbf{98.2}\\
019\_pitcher\_base  & \textbf{96.7} & 97.8 & 52.9 & 77.6 & 85.9 & 93.6 & 77.6 & 91.7 & 92.1 & \underline{98.0} & 94.0 & 97.3 & \textbf{96.7} & \textbf{98.7}\\
021\_bleach\_cleanser  & \textbf{92.3} & \underline{96.8} & 50.2 & 71.7 & 74.7 & 87.6 & 42.0 & 62.6 & 54.4 & 70.9 & 82.1 & 92.7 & \underline{90.0} & \textbf{97.8}\\
024\_bowl  & 17.5 & 78.3 & 3.0 & 69.6 & 5.5 & 78.0 & 17.3 & 79.1 & 19.6 & 79.5 & \underline{34.5} & \underline{87.8} & \textbf{36.6} & \textbf{89.8}\\
025\_mug  & 81.4 & 95.1 & 58.4 & 78.8 & \underline{88.2} & \underline{96.6} & 58.0 & 80.9 & 82.8 & 93.6 & 87.8 & 94.9 & \textbf{94.9} & \textbf{98.2}\\
035\_power\_drill  & \textbf{96.9} & \underline{98.0} & 55.2 & 73.2 & 95.1 & 97.9 & 61.2 & 77.9 & 81.9 & 89.3 & 89.7 & 95.1 & \underline{96.2} & \textbf{98.3}\\
036\_wood\_block  & \textbf{79.2} & \textbf{90.5} & 26.4 & 64.3 & \underline{35.5} & 69.9 & 1.6 & 55.2 & 2.5 & 60.8 & 24.8 & 80.5 & 29.0 & \underline{87.4}\\
037\_scissors  & \textbf{78.4} & \textbf{92.2} & 34.8 & 55.9 & 59.0 & 79.6 & 0.8 & 7.0 & 0.7 & 7.5 & 67.6 & 81.5 & \underline{73.9} & \underline{86.9}\\
040\_large\_marker  & \underline{85.4} & \underline{97.2} & 58.2 & 71.9 & 83.6 & 95.3 & 55.6 & 67.6 & 65.9 & 75.7 & 84.3 & 93.1 & \textbf{90.8} & \textbf{97.5}\\
051\_large\_clamp  & \textbf{52.6} & 75.4 & 24.6 & 50.1 & \underline{50.7} & 74.0 & 32.8 & 72.2 & 41.2 & 83.3 & 40.1 & \underline{91.3} & 40.8 & \textbf{94.6}\\
052\_extra\_large\_clamp  & 28.7 & 65.3 & 16.3 & 44.5 & 25.8 & 67.7 & 26.9 & 59.5 & 32.5 & 63.6 & \underline{40.2} & \textbf{75.7} & \textbf{40.5} & \underline{75.1}\\
061\_foam\_brick  & 48.3 & \underline{97.1} & 40.4 & 88.2 & 42.2 & 92.5 & 19.4 & 56.2 & 22.5 & 57.7 & \underline{51.7} & 94.7 & \textbf{52.7} & \textbf{97.6}\\
\noalign{\smallskip}
\hline
\noalign{\smallskip}
\textbf{All Frames}  & \textbf{79.3} & \underline{93.0} & 53.7 & 76.3 & 73.1 & 89.3 & 50.5 & 72.8 & 61.2 & 80.3 & 76.1 & 92.6 & \underline{78.9} & \textbf{94.7}\\
\noalign{\smallskip}
\hline
\end{tabularx}
\end{table*}%
\footnotetext[1]{\url{https://github.com/yuxng/YCB\_Video\_toolbox}}\footnotetext[2]{\url{https://github.com/DLR-RM/AugmentedAutoencoder}}\footnotetext[3]{\url{https://github.com/ylabbe/cosypose}}%
The comparison demonstrates that \textit{ICG} is able to outperform most methods by a considerable margin for the \textit{ADD-S} score, performing on par with the best reported results from \textit{FFB6D}.
This is remarkable since \textit{FFB6D} trains on large amounts of pose-annotated real-world data that originates from a similar pose distribution.
For many applications, such data is not available.
In contrast, \textit{ICG} only requires a texture-less 3D model and no training data.
With respect to the \textit{ADD(S)} metric, both \textit{PVN3D} and \textit{FFB6D} report better results.
The main reason for this is that our method is by design not considering texture and thus has a considerable disadvantage if the geometry is not conclusive.
However, in return, there is no need for textured 3D models, which are required for all competing methods.

In contrast to \acs{6DoF} pose estimation algorithms, \textit{ICG} considers the pose on a frame-to-frame basis without re-initialization.
This leads to a small number of objects that get stuck in local minima and thus show relatively poor performance.
On the other hand, \textit{ICG} benefits from temporal consistency and performs more accurate than \textit{FFB6D} for most objects.
The advantage of our tracker becomes particularly obvious when comparing the required computation.
While \textit{FFB6D} depends on a high-end GPU and reports a runtime of $75\,\unit{ms}$ \cite{He2021}, \textit{ICG} requires only $1.3\,\unit{ms}$ per frame on a single CPU core, which is $57\times$ faster.
This is especially crucial in reactive, real-time applications for which hardware constraints exist.
In conclusion, the experiment demonstrates that while tracking by detection is possible, for many real-world applications, it is not the most sensible solution.
Given the high efficiency and good performance of \textit{ICG}, in our opinion, it is best to rely on continuous 3D tracking for local pose updates while using \acs{6DoF} pose estimation for global initialization and long-term consistency.

\section{6-DoF Pose Refinement}\label{sec:a4}
Given that \textit{ICG} is a local optimization method, the question emerges how well it would work for pose refinement.
In the following, we thus use \textit{ICG} to improve the predictions of \textit{PoseCNN}, \textit{Augmented Autoencoders}, and \textit{CosyPose} and compare results on the \textit{YCB-Video} dataset \cite{Xiang2018}.
Depending on the pose estimation algorithm, errors along the principal axis are relatively large.
To cope with those larger translational errors, we use the following parameters $r_\textrm{t} = \{300, 250, 100\}$, $\sigma_\textrm{d} = \{100, 50, 20\}$, $\lambda_\textrm{t} = 100$, and conduct $7$ instead of $4$ iterations.
For efficiency, strides are increased from $5\,\unit{mm}$ to $10\,\unit{mm}$.
All other parameters remain the same as in the evaluation of tracking.
With the increased number of iterations and considered area, the runtime increases to $2.7\,\unit{ms}$ per frame.

Results of the conducted evaluation are shown in \cref{tab:a03}.
We thereby report both refined and unrefined scores for the considered \acs{6DoF} pose estimation methods.
In addition, results from \cite{Xiang2018} are provided, which were obtained using an extensive multi-hypothesis \textit{ICP} approach on the predictions of \textit{PoseCNN}.
According to \cite{Wang2019b}, this refinement algorithm requires more than $10\,\unit{s}$ for a single pose.
The evaluation shows that, even for the very good results of \textit{CosyPose}, \textit{ICG} is able to improve pose estimations for almost all objects.
Also, it is interesting to see that, while it can not fully compete with extensive multi-hypothesis \textit{ICP} refinement, the difference is not as big as one might expect.
This is especially impressive considering that \textit{ICG} is more than three orders of magnitude faster.

Finally, we want to ensure that the pose refinement uses both depth and region information and that improvements are not only from the \textit{ICP}-based depth modality.
We thus conducted a short ablation study, for which results are shown in \cref{tab:a04}.
\begin{table}
	\caption{
		Ablation study comparing refined results for \textit{ICG} with and without the region modality to unrefined results.
		Values show \textit{ADD} and \mbox{\textit{ADD-S}} area under curve scores over all frames on the \textit{YCB-Video} dataset \cite{Xiang2018} in percent.
	}\label{tab:a04}

\centering
\scriptsize
\begin{tabularx}{\linewidth}{X@{\hspace{0.0cm}} | >{\centering\arraybackslash}p{0.79cm}@{\hspace{0.0cm}} >{\centering\arraybackslash}p{0.79cm} | >{\centering\arraybackslash}p{0.97cm}@{\hspace{0.0cm}} >{\centering\arraybackslash}p{0.97cm} | >{\centering\arraybackslash}p{0.79cm}@{\hspace{0.0cm}} >{\centering\arraybackslash}p{0.79cm}}
\hline
\noalign{\smallskip}
\multirow{2}{1cm}{\textbf{Approach}} & 
\multicolumn{2}{c|}{\multirow{2}{1.58cm}{\centering PoseCNN\protect\footnotemark[1] \cite{Xiang2018}}}&
\multicolumn{2}{c|}{\multirow{2}{1.94cm}{\centering Augmented Autoencoders\protect\footnotemark[2] \cite{Sundermeyer2018}}}&
\multicolumn{2}{c}{\multirow{2}{1.58cm}{\centering CosyPose\protect\footnotemark[3] \cite{Labbe2020}}}\\
&\multicolumn{2}{c|}{}&
\multicolumn{2}{c|}{}&
\multicolumn{2}{c}{}\\
\noalign{\smallskip}
\hline
\noalign{\smallskip}
Refinement & ADD & ADD-S & ADD & ADD-S & ADD & ADD-S \\
\noalign{\smallskip}
\hline
\noalign{\smallskip}
Unrefined & 53.7 & 76.3 & 50.5 & 72.8 & 76.1 & 92.6\\
ICG w/o Region & 65.0 & 84.2 & 57.5 & 76.9 & 76.8 & 93.3\\
ICG w/ Region& 73.1 & 89.3 & 61.2 & 80.3 & 78.9 & 94.7\\
\noalign{\smallskip}
\hline
\end{tabularx}
\end{table}
The obtained scores demonstrate that \textit{ICG} is not just a blown-up \textit{ICP} approach but that the addition of region information significantly helps to improve performance.
Given the good pose predictions and computational efficiency, we are thus confident that, while \textit{ICG} is an excellent 3D object tracking approach, it also has many applications in pose refinement.

\end{document}